\documentclass[final,1p,times]{elsarticle}
\usepackage[T1]{fontenc}
\usepackage{csvsimple}
\usepackage{adjustbox,lipsum}
\usepackage{amsmath,amssymb,amsfonts}
\usepackage{graphicx}
\usepackage{hyperref}
\usepackage{url}
\usepackage{newfloat}
\usepackage{ulem}
\usepackage{multirow}
\usepackage{algorithm}
\usepackage{enumitem}
\usepackage{xcolor}
\usepackage{comment}
\usepackage{caption}
\usepackage{subcaption}
\usepackage{algorithmicx}
\usepackage{algpseudocode}
\usepackage{cases}
\usepackage{tabularx}
\usepackage{nicefrac}       %
\usepackage{amsthm}
\usepackage{microtype}      %
\usepackage{thmtools}

\restylefloat{figure}

\usepackage[colorinlistoftodos]{todonotes} %
\definecolor{lightgray}{HTML}{E8E8E8}

\journal{Artificial Intelligence}

\usepackage{dsfont}
\usepackage{tikz}
\usepackage{pgfplots,pgfplotstable,booktabs}
\usepgfplotslibrary{fillbetween}
\usetikzlibrary{patterns}
\usetikzlibrary{shapes}
\usetikzlibrary{plotmarks}

\pgfmathdeclarefunction{lognorm}{1}{%
  \pgfmathparse{1/(#1*x*sqrt(2*pi))*exp(-(ln(x))^2/(2*#1^2))}%
}

\DeclareMathOperator*{\argmax}{arg\,max}
\newcommand{\set}[1]{\left\{ \left. #1 \right. \right\}}

\declaretheorem{theorem} 
\declaretheoremstyle[%
  spaceabove=-6pt,%
  spacebelow=6pt,%
 headfont=\normalfont\itshape,%
  postheadspace=1em,%
  qed=\qedsymbol]{mystyle} 
\declaretheorem[name={Proof},style=mystyle,unnumbered]{prf}
\theoremstyle{definition}
\newtheorem{definition}{Definition}

\theoremstyle{remark}
\newtheorem*{remark}{Remark}

\newcommand{\sachant}{\, \right| \left. \,}
 
\newcommand{\croch}[1]{\left[ \left. #1 \right. \right]}
\definecolor{britishracinggreen}{rgb}{0.0, 0.26, 0.15}

\newcommand{\eref}[1]{Eq. (\ref{#1})}
\newcommand{\defeq}{\stackrel{\mathrm{def}}{=}}

\pgfplotsset{compat=1.17}

\definecolor{bblue}{HTML}{4F81BD}
\definecolor{rred}{HTML}{C0504D}
\definecolor{ggreen}{HTML}{9BBB59}
\definecolor{ppurple}{HTML}{9F4C7C}

\usetikzlibrary{positioning}
\usetikzlibrary{decorations.text}

\begin{document}

\begin{frontmatter}

\title{
An Offline Risk-aware Policy Selection Method for Bayesian Markov Decision Processes}

\author[addressaniti,addressisae]{Giorgio Angelotti}
\corref{mycorrespondingauthor}
\cortext[mycorrespondingauthor]{Corresponding author}%

\address[addressaniti]{ANITI - Artificial and Natural Intelligence Toulouse Institute, University of Toulouse, France}
\address[addressisae]{ISAE-SUPAERO, University of Toulouse, France}

\author[addressaniti,addressisae]{Nicolas Drougard}

\author[addressaniti,addressisae]{Caroline P.~C.~ Chanel}
 
\begin{abstract}
In Offline Model Learning for Planning and in Offline Reinforcement Learning, the limited data set hinders the estimate of the Value function of the relative Markov Decision Process (MDP). Consequently, the performance of the obtained policy in the real world is bounded and possibly risky, especially when the deployment of a wrong policy can lead to catastrophic consequences. For this reason, several pathways are being followed with the scope of reducing the model error (or the distributional shift between the learned model and the true one) and, more broadly, obtaining risk-aware solutions with respect to model uncertainty.
But when it comes to the final application which baseline should a practitioner choose?
In an offline context where computational time is not an issue and robustness is the priority we propose Exploitation vs Caution (EvC), a paradigm that (1) elegantly incorporates model uncertainty abiding by the Bayesian formalism, and (2) selects the policy that maximizes a risk-aware objective over the Bayesian posterior between a fixed set of candidate policies provided, for instance, by the current baselines.
We validate EvC with state-of-the-art approaches in different discrete, yet simple, environments offering a fair variety of MDP classes. In the tested scenarios EvC manages to select robust policies and hence stands out as a useful tool for practitioners that aim to apply offline planning and reinforcement learning solvers in the real world.
\end{abstract}

\begin{keyword}
Risk-aware Markov Decision Process
\sep Bayesian Markov Decision Process
\sep Offline Policy Selection
\sep Offline Model Learning for Planning
\sep Offline Reinforcement Learning

\end{keyword}
\end{frontmatter}

\section{Introduction}
\noindent The deployment of autonomous agents in an unknown, and possibly stochastic, environment is a delicate task that usually requires a continuous agent-environment interaction %
which is not always affordable in real-life situations. For instance, in applications such as the training of medical robots and automated vehicles \cite{medical,selfdrivingcars} the interaction with the environment can be both too risky (e.g. proximity to a human) and costly since: (i) any mistake could lead to catastrophic aftermaths; or (ii) the data collection phase requires a direct human involvement, which is usually expensive and time demanding. Hence, it can be convenient to exploit previously collected data sets in order to limit additional (dangerous or superfluous) interaction.\\
Offline model learning for planning and offline Reinforcement Learning (RL) are the branches of machine learning that leverage previously collected batches of experiences with the aim of establishing an optimal behavioral policy offline. In recent years, the RL community published a great number of papers on the subject, as for instance, the works in \cite{spibb, BCQ2,BEAR,BRAC,bopah,Chen,MOPO,MOReL,Levine}, demonstrating the growing interest in the field. %
The proposed algorithms try to improve the performance of a policy obtained either with model-free or model-based RL approaches. The intuition behind these methods is always the same: optimizing a trade-off between \textit{exploitation and caution}. The policy optimization procedure is usually tailored in order to generate strategies that are not too distant from the one originally used to collect the batch. In this way, the agent will follow a strategy that will not drive him towards regions of the state-action space for which it possesses a high degree of uncertainty. 

For instance, in the offline RL literature tailored policy optimization procedures for Markov Decision Processes (MDPs) are implemented:
(i) in Conservative Q-Learning (CQL) approach \cite{kumar2020conservative}, by limiting the overestimation of Out-Of-Distribution transitions;
(ii) in the Behavior Regularized Actor-Critic (BRAC) paradigm \cite{BRAC}, as a penalty in the value function proportional to an estimate of the policies' distributional shift;
(iii) in the Model-based Offline Policy Optimization (MOPO) algorithm \cite{MOPO}, as a penalization added to the reward function which is proportional to an estimate of the distributional shift in the dynamical evolution of the system - also called \textit{epistemic} (model) error;
and, (iv) in the Model-Based Offline
Reinforcement Learning (MOReL) approach \cite{MOReL}, by creating an additional and highly penalized absorbing state and by forcing the agent to transit to it when the model uncertainty for a specific state-action pair is above a given threshold.

On top of a non-trivial estimate of per-transition uncertainty, which is often performed with Deep Neural Networks, the said baselines notably require the fine-tuning of domain-dependent hyperparameters \cite{paine2020hyperparameter}. Such an empirical calibration %
demands additional interaction with the environment and thus betrays the original purpose of offline learning.

How to select the best set of hyperparameters for offline RL baselines based on Deep Neural Networks?
This question does not have a trivial answer \cite{Levine}. Usually, the policies obtained are evaluated offline with \textit{off policy-evaluation} (OPE) approaches like Fitted Value Iteration \cite{munos2008finite} or Fitted Q Evaluation (FQE) \cite{le2019batch}. Indeed, \cite{paine2020hyperparameter} showed that ORL baselines using DNN are not robust with respect to hyperparameter selection and presented a comparison between offline hyperparameter selection methods. Nevertheless, algorithms like FQE require hyperparameters to be tuned, shifting then the problem of selecting the best hyperparameters for the ORL baseline to the one of selecting the hyperparameters for the OPE. Recently, a totally hyperparameter-free method called BVFT for OPE has been proposed in the work in \cite{zhang2021towards}. However, the said method is affected by limited data efficiency and computational complexity that scales quadratically with the number of models to compare. Concurrently, the work in \cite{yang2021pessimistic} presented a pessimistic method to estimate and select models for Offline Deep RL.

From another perspective, the Safe Policy Improvement with Baseline Bootstrapping (SPIBB) methodology \cite{spibb} and the Batch Optimization of Policy and Hyperparameter (BOPAH) approach \cite{bopah} were developed with the aim of obtaining a robust policy with a true hyperparameter agnostic approach. The former generates a safe policy improvement over the data collecting policy with theoretical guarantees similar to the ones achievable in Probably Approximately Correct approaches, the latter uses a classic batch (offline) RL approach with a gradient-based optimization of the hyperparameters using held-out data. %

Aside from the RL community, researchers whose field is mostly offline model learning for planning deal with the problem of solving MDPs under model uncertainty by focusing on the resolution of \textit{robust MDPs} \cite{nilim2005robust, iyengar2005robust}. In this context, the model dynamics (e.g. a stochastic transition function) lie in a constrained ambiguity set which is a subset included in the whole set of distributions. The problem is hence formulated as a dynamic game against a malevolent nature which at every time step chooses the worst model in the set according to the agent action. 
Subsequently to these works, the research in \cite{delage2010percentile} introduced the \textit{chance constrained MDP} approach which optimizes policies for the \textit{percentile criterion}: the Value-at-Risk (\textit{VaR}) metric. Reference \cite{delage2010percentile} proved that robust MDPs can generate overly conservative strategies depending on the size and shape of the ambiguity set. And, reference \cite{ghavamzadeh2016safe} proposed approximate solutions to generate safe policy improvements of the data collector policy.
Recently, the works in \cite{petrik2019beyond, pmlr-v130-behzadian21a} incorporated prior knowledge upon a Bayesian methodology in order to obtain less conservative ambiguity sets that can yield tighter safe returns estimates. In particular, the study in \cite{petrik2019beyond}
proposes \textit{Bayesian Credible Region} (BCR), an algorithm that constructs ambiguity sets from Bayesian credible (or confidence) regions and uses them to optimize the risk-aware problem (the robust MDP).
Reference \cite{lobo2021softrobust} introduced the \textit{Soft-Robust Value Iteration} (SRVI) algorithm to optimize for the \textit{soft-robust criterion}, a weighted average between the classic value function and the Conditional \textit{VaR} risk metric with epistemic model uncertainty, to solve a Robust MDP. %

In parallel to these works, reference \cite{gamma} showed that planning in an MDP context using a discount factor $\gamma^*$ lower than the one used in the final evaluation phase $\gamma_{ev}$ yields more performing policies when a trivially learned MDP model is considered. A trivially learned MDP model is said to be the one that maximizes the likelihood of the transitions collected in the batch.
Nevertheless, the mathematical expression that should be optimized in order to find $\gamma^*$ is intractable. In the work in \cite{gamma}, the optimal discount factor is finally found by cross-validation which requires additional interaction with the true environment.

Off-policy evaluation for finite, discrete MDPs usually resorts to techniques based on Importance Sampling \cite{precup2000eligibility, Levine}. The Importance Sampling procedure assigns different weights to samples when one exploits them to estimate values from a distribution that is different from the one that was used to generate the samples. This weight, called the Importance Sampling ratio, in off-policy evaluation is the probability of sampling a specific trajectory using the new policy over the probability of obtaining the same trajectory using the policy used to collect the data. This ratio is independent of the models' transition function and can be simplified as the ratio between the probabilities of generating that given sequence of actions while deploying the two different policies. The Universal Off-Policy Evaluation (UnO) \cite{chandak2021universal} has been proposed to estimate not only the average value and the variance of policy performance but also risk-sensitive metrics based on quantiles like the \textit{VaR} or the \textit{CVaR}. UnO is a non-parametric and model-free estimator based on Importance Sampling for the cumulative distribution of returns of a fixed policy $\pi$ starting from a pre-collected batch of experiences. Estimating the full cumulative distribution allows computing risk-sensitive metrics like the Value at Risk and the Conditional Value at Risk.

Unfortunately, UnO and other Importance Sampling based techniques manage to properly estimate the MDP's value function only for stochastic policies while many policies generated by state-of-the-art approaches are deterministic. On one hand, it is true that a deterministic policy is just a specific type of stochastic policy, but on the other hand, the computation of the importance sampling ratio for deterministic policies collapses onto a Kronecker delta. Eventually, the offline evaluation of deterministic policies with Importance Sampling ratio is not accurate. There is a necessity of developing a technique to evaluate and select offline robust deterministic  policies.

Concurrently, with the aim of finding a policy that optimizes the trade-off between \textit{exploitation and exploration} in an \textit{online} setting, model uncertainty has been included in a Bayesian extension of the MDP framework called Bayesian (Adaptive) MDP (BAMDP) \cite{strens}. Fixing a prior for the distribution over transition models, a posterior distribution is computed from the likelihood of the sampled trajectories. %
Some years later, the work in \cite{Sharma} suggested that risk-aware utility functions can replace the common BAMDP value function. In doing so, the said work proposed an algorithm that trades off exploration, exploitation, and robustness (caution). The said works (e.g. \cite{strens,Sharma}) deal with an online context when interaction with the environment is always possible while, as we stated before in this manuscript, we will tackle an offline problem.
The works in \citep{depeweg2017uncertainty,depeweg2018decomposition} exploit Bayesian Neural Networks with latent variables to encode the uncertainty in model-based Reinforcement Learning with the generator model represented as a Deep Neural Network. On top of this, optimization for a risk-sensitive utility function that is the future expected return plus a variance term that includes both noise and model uncertainty is performed. Unfortunately, the said risk-sensitive objective can unlikely mitigate efficiently the risk in environments where the distribution of returns is far from Gaussian, \textit{e.g.} multi-modal distributions.

Solving a Bayesian MDP is NP-hard \cite{steimle2021multi}. Moreover, the space $\mathbb{M}$ over which the posterior distribution of models is defined is an infinite set. To overcome this computational constraint, the work in \citep{steimle2021multi} proposes to find the policy that maximizes a utility criterion over a finite list of models. The said framework is called Multi-model MDP. While the maximization of the utility criterion using a finite list of models is more treatable than solving a Bayesian MDP, constraining a possibly infinite set of models to a set of candidate ones could in many cases be a too strong approximation. In this work, we will not limit the space of possible models, but instead of solving a Bayesian MDP, we will rather  focus on the problem of selecting the most performant policy from a set of candidates, according to a risk-sensitive criterion, over the full distribution of possible models.

Interestingly, the problem of selecting the right algorithm for a machine learning problem was studied in the work in \citep{rice1976algorithm} that formalized the protocol to be followed to analyze and solve an algorithm selection problem. And, following the work in \cite{rice1976algorithm}, Upper Confidence Bound (UCB-1) for Multi-Armed Bandits has been applied in \cite{laroche2017reinforcement} for algorithm (and hence policy) selection in Reinforcement Learning, however, the said approach was not risk-aware and was limited to an online setting.

In fact, in light of the current limitations of state-of-the-art, we notice that selecting a robust policy for an offline data-driven MDP taking into account the uncertainty in the model learning phase is still an open problem. In this context, our work presents Exploitation vs Caution (EvC), an \textit{offline} Bayesian paradigm to evaluate the performance of the policies provided by the state-of-the-art solvers and \textit{select} the best policy between a set of candidates according to different risk-metrics (\textit{VaR} and also the Conditional \textit{VaR}). The set of candidate policies is initially obtained using the available baselines. Taking inspiration from the works in \cite{gamma} and \cite{Sharma}, the set of candidate policies is then enriched considering strategies that are obtained by solving several MDPs with different discount factors and different transition functions which are sampled from the Bayesian posterior inferred from the original fixed batch of pre-collected transitions.
The distribution of the performance of every policy taking into account the model uncertainty is evaluated by alternating Monte Carlo model sampling and Policy Evaluation. In the end the best policy according to the risk-sensitive criterion is selected.
\newpage

\noindent In summary, the contributions of this work are:
\begin{enumerate}
    \item it proposes a paradigm to select the best risk-aware policy between a set of candidate policies for a Bayesian MDP in the offline setting;
    \item it gives probabilistic guarantees on the performance of the selected policy by computing risk-aware criteria through a confident estimation method for a given risk level (or quantile order);
    \item the empirical results demonstrate the validity of this approach in toy environments compared to the current policy selection baseline (UnO).
    
\end{enumerate}
The paper is organized as follows:
Section \ref{sec:mdp} starts with
a recap of the MDP and of the Bayesian MDP (BMDP) formalisms; 
risk-aware measures following the prescriptions of the research in \cite{risk} and the Risk-aware BMDP are introduced in Section \ref{sec:measures}; in Section \ref{sec:evc} the EvC approach is presented; 
then in Section \ref{sec:exp}, the policy selected by EvC is compared against selected baseline approaches, and to a risk-sensitive policy selection approach achieved by evaluating the set of candidate policies with UnO;
Section \ref{sec:conclusions} concludes the paper by discussing its limitations %
and pointing to future work perspectives.
\section{Background}
\label{sec:mdp}
\begin{definition}%
\label{def:mdp}
\it{A Markov Decision Process (MDP) is a $6$-tuple $M \defeq \langle \mathcal{S}, \mathcal{A}, T, r, \gamma, \mu_0\rangle$
where $\mathcal{S}$ is the set of states, $\mathcal{A}$ the set of actions, $T:  \mathcal{S}\times\mathcal{A}\times \mathcal{S}\rightarrow \left[0, 1\right]$ is the state transition function $T(s,a,s')$ defining the probability that dictates the evolution from $s \in \mathcal{S}$ to $s' \in \mathcal{S}$ after taking the action $a \in \mathcal{A}$, $r: \mathcal{S} \times \mathcal{A} \rightarrow \left[ r_{\text{min}}, r_{\text{max}}\right]$, with $r_{\text{max}}, r_{\text{min}} \in \mathbb{R}$, is the reward function $r(s,a)$ that indicates what the agent gains when the system state is $s\in \mathcal{S}$ and action $a \in \mathcal{A}$ is applied, $\gamma \in \left[0, 1\right)$ is called the discount factor and $\mu_0 : \mathcal{S} \rightarrow \left[0,1\right]$ is the distribution over initial states: $\sum_{s \in \mathcal{S}}\mu_0(s)=1$.
}
\end{definition}
\begin{definition}%
\it{A policy is a %
mapping from states to a probability distribution over actions, such as $\pi: \mathcal{A}\times \mathcal{S}\rightarrow \left[0, 1\right]$.}
\end{definition}
\begin{definition}%
\it{Solving an MDP amounts to finding a policy $\pi^*$ which, $\forall s \in \mathcal{S}$, maximizes the value function:
\noindent
\begin{equation}
V_M^\pi\left(s\right) \defeq \mathbb{E}_{\substack{A_t \sim \pi\\ S_{t} \sim T}}\croch{\sum_{t=0}^{\infty}\gamma^t r\left(S_t, A_t\right) \sachant S_0 = s}. \label{eq:mdpvaluefunction}\end{equation}}
\end{definition}
It has been proved that an MDP for which the value function is defined as \eref{eq:mdpvaluefunction} admits a deterministic optimal policy (a map from states to actions) \cite{kolobov2012planning}:
\begin{equation}
    \pi^*\left(s\right) = \text{argmax}_{\pi} V_{M}^{\pi}(s).%
\end{equation}
\begin{definition}
\it{
The performance of a policy $\pi$ in an MDP $M$ %
with value function $V_{M}^{\pi}$ is defined as:
\begin{equation}
\label{eq:performance}
    u^{\pi}(M) = \mathbb{E}_{S \sim \mu_{0}}\big[V_{M}^{\pi}(S)\big].
\end{equation}}
\end{definition}
\begin{definition}%
\label{def:bmdp}
\it{A BMDP is a $8$-tuple $\beta \defeq \langle \mathcal{S}, \mathcal{A}, \mathcal{T}, \mathcal{R}, \tau, \rho, \gamma, \mathcal{B} \rangle$ where $\mathcal{S}$ is the set of states; $\mathcal{A}$ the set of actions; $\mathcal{T}$ is a parametric family of transition functions $T$ of any MDP compatible with $\mathcal{S}$ and $\mathcal{A}$: $\mathcal{T} = \big\{ T: \mathcal{S} \times \mathcal{A} \times \mathcal{S} \rightarrow [0,1] \mbox{ s.t. } \sum_{s' \in \mathcal{S}} T(s,a,s') =1 \big\}$}; $\mathcal{R}$ is a parametric family of reward functions $r$ of any MDP compatible with $\mathcal{S}$ and $\mathcal{A}$: $\mathcal{R} = \big\{ r: \mathcal{S} \times \mathcal{A} \rightarrow [r_{min}, r_{max}] \big\}$; $\tau$ is a non-informative prior distribution uniform over $\mathcal{T}$: $\int_{T \in \mathcal{T}}d\tau=1$ with $\tau\geq0$; $\rho$ is a non-informative prior distribution uniform over $\mathcal{R}$: $\int_{r \in \mathcal{R}}d\rho=1$ with $\rho\geq0$; $\gamma \in \left[0,1\right)$ is the discount factor, and $\mathcal{B} = \left\{\left(s_t, a_t, r_{t}, s_{t+1}\right)\right\}$ is a batch of transitions %
generated by acting in a fixed, unknown MDP compatible with $\mathcal{S}$ and $\mathcal{A}$ %
and initial state distribution $\mu_0$.
\end{definition}
For instance, in a finite state and action spaces environment, $\mathcal{T}$ is the set of all $\lvert \mathcal{S}\rvert \times \lvert \mathcal{A}\rvert$ different discrete distributions and $\tau$ is made of $\lvert \mathcal{S}\rvert \times \lvert \mathcal{A}\rvert$ uniform (uninformative) Dirichlet probability density functions -- the conjugate prior of the said distribution.
\begin{definition}
\it{$\tau_p$ is a posterior distribution over $\mathcal{T}$ obtained by updating the uniform (uninformative) Dirichlet prior $\tau$ with the information contained in $\mathcal{B}$.}
\end{definition}
In particular, the $\lvert \mathcal{S}\rvert$ probability values $X_i$ with $i \in \{1,
\dots, \mathcal{S}\}$ describing the probability of %
$\left(S=s^*, A=a^*\right)\rightarrow (S'=s_i)$ can be distributed as:
\noindent
\begin{equation}
\label{eq:post}
\tau_p^{s^{*},a^{*}}\left(x_1, \dots, x_{\lvert \mathcal{S}\rvert} \big\lvert n_1, \dots, n_{\lvert \mathcal{S}\rvert}\right) = \Gamma\left(\nu\right)\prod_{i=1}^{\lvert \mathcal{S}\rvert}\dfrac{x_{i}^{n_{i}}}{\Gamma\left(n_{i}+1\right)}
\end{equation}
where, $\Gamma$ is the Euler gamma function, $n_i$ counts how many times the transition $\left(s^*,a^*\right)\rightarrow s_i$ appears in $\mathcal{B}$ and $\nu = \sum_{k=1}^{\lvert S\rvert}\left(n_{k} +1\right)$.
\begin{remark}
Notice that the mode (the most likely configuration) of the posterior in \eref{eq:post} is given by $\hat{x}_i=\frac{n_i}{\sum_{k=1}^{\lvert S \rvert}n_{k}}$ while its expected value is $\mathbb{E}_{\tau_p}\left[X_i\right]=\frac{n_i+1}{\nu}$. %
\end{remark}
Since we consider discrete environments, the most likely transition model with respect to $\tau_p$ is the one for which the transition probabilities are given by the transition frequencies in $\mathcal{B}$. We refer to these distributions as the trivial model, noted as $\hat{T}$.
We note that a similar reasoning can be applied to $\mathcal{R}$ and $\rho$, however for simplicity's sake %
we assume to know the reward function $r$ in this work.

\begin{remark}
It would be possible to define a prior over the initial states and obtain a posterior taking into account the information contained in the batch $\mathcal{B}$. For simplicity, we will %
also assume that $\mu_0$ is known.
\end{remark}
\begin{definition}%
\it{A solution to a BMDP $\beta$ is a policy $\pi$ which maximizes the following utility function:
\begin{equation}
\label{eq:bayesian}
    \mathcal{U}^{\pi}_{\beta} \defeq \mathbb{E}_{M \sim \tau_p}\left[ u^{\pi}(M)\right]
\end{equation}
where, $u^{\pi}(M) \defeq \mathbb{E}_{S\sim\mu_0}\left[V_M^{\pi}\left(S\right)\right]$ is the expected value of an MDP, averaged on the initial state, with transition function sampled from $\tau_p$.}
\end{definition}

The optimal performance with respect to \eref{eq:bayesian} will be the one that, on average, works the best on the BMDP $\beta$ when the model is distributed according to the Bayesian posterior:
\begin{equation}
\label{eq:npobj}
\mathcal{U}^{*}_{\beta} = \displaystyle \max_{\pi}
\mathcal{U}^{\pi}_{\beta}
\end{equation}

\begin{remark}
Since the true MDP is unknown, leveraging the Bayesian framework is an elegant way to incorporate %
uncertainty. However, the additional expected value makes \eref{eq:bayesian} hard to be computed with Bellman's recursive approaches or approximated with temporal differences methods. Indeed, the objective stated in \eref{eq:npobj} is NP-hard \cite{steimle2021multi}.
\end{remark}

\section{Risk-aware measures}
\label{sec:measures}
We advocate that solving a BMDP deals with uncertainty more elegantly than solving an MDP for the most likely model $\hat{M}=(\hat{T},\hat{r})$. However, the utility function defined in \eref{eq:bayesian} does not minimize the risk of obtaining a bad performant policy in the real environment. 

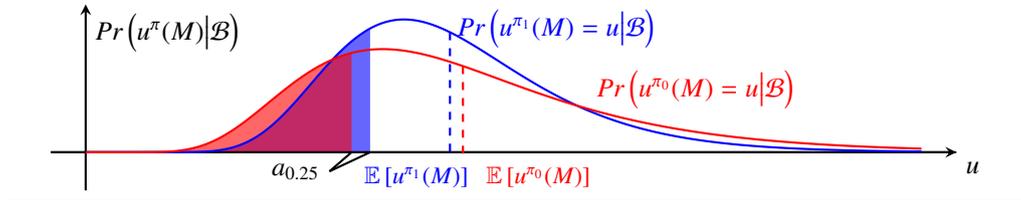
\begin{figure}[!ht]
\centering
\begin{tikzpicture}
\def\mufour{1.08328707}
\def\muthree{1.04602786}
\def\fourtall{0.902}
\def\threetall{1.257}
\def\xMin{-0.1}
\def\xMax{2.5}
\def\yMax{1.55}
\def\eps{0.01}
\def\xVarq{0.7}
\def\yVarq{-0.2}
\begin{axis}[xmin=\xMin, xmax=\xMax,
	ymin=-0.4, ymax=\yMax,
	ticks=none,
	axis line style=thick,
    every axis plot post/.append style={
	mark=none,
	domain=\xMin:\xMax-0.1,
	samples=1000,
	thick
	},
width=\textwidth,
height=0.3\textwidth,
xlabel={$u$}, 
ylabel={$Pr\left(u^{\pi}(M)\big\lvert \mathcal{B}\right)$},
axis x line=middle,
axis y line=middle,
every axis x label/.style={
    at={(current axis.right of origin)},
    anchor=north west,
}
]
\definecolor{rb}{rgb}{0.5, 0.0, 0.5}
\addplot[name path=plot1] coordinates {(\xMin,0)(\xMax,0)};
\addplot[white] coordinates {(\xMax,1.8)};
\addplot[name path=plot2, blue]{lognorm(0.3)};
\addplot[name path=plot3, red]{lognorm(0.4)};
\addplot [fill=blue, fill opacity=0.6] fill between[of= plot1 and plot2, soft clip={domain=\xMin:0.8168115}];
\addplot [fill=red, fill opacity=0.6] fill between[of= plot1 and plot3, soft clip={domain=\xMin:0.7635353159}];
\addplot[blue, nodes near coords*={$Pr\left(u^{\pi_1}(M)=u\big\lvert\mathcal{B}\right)$}] coordinates {(1.35,\yMax-0.56)};
\addplot[red, nodes near coords*={$Pr\left(u^{\pi_0}(M)=u\big\lvert\mathcal{B}\right)$}] coordinates {(1.75,\yMax-1.2)};
\addplot[blue,dashed] coordinates {(\muthree,\threetall) (\muthree,-0.04)}; %
\addplot[red,dashed] coordinates {(\mufour,\fourtall)(\mufour,-0.07)};
\addplot[red, nodes near coords*={\small$\mathbb{E}\left[u^{\pi_0}(M)\right]$}] coordinates {(1.3,\yMax-1.6)};
\addplot[blue, nodes near coords*={\small$\mathbb{E}\left[u^{\pi_1}(M)\right]$}] coordinates {(0.95,\yMax-1.6)};
\addplot[black] coordinates {(0.8168115,0)(\xVarq,\yVarq)};
\addplot[black] coordinates {(0.7635353159,0)(\xVarq,\yVarq)} node[anchor=east] {$a_{0.25}$};
\end{axis}
\end{tikzpicture}
\caption{\label{fig:meanvar} Pdf for the performance $u^{\pi}(M)=u$ given the batch $\mathcal{B}$ and two different policies $\pi_0$ and $\pi_1$. Dashed lines correspond to the expected value. Both curves are filled up to their $0.25$-quantiles ($a_{0.25}$). The red curve has a higher expected value while the blue one corresponds to a safer policy.}%
\end{figure}

For instance, let $Pr\left(u^{\pi}(M)=u|\mathcal{B}\right)$
be the probability density function (pdf) of the performance of the policy $\pi$ when the model is distributed according to the Bayesian posterior $\tau_p$. Given two policies $\pi_0$ and $\pi_1$, we can have that $\mathbb{E}\left[u^{\pi_0}(M)\right]>\mathbb{E}\left[u^{\pi_1}(M)\right]$ (see Figure \ref{fig:meanvar}). In this case, following the BMDP optimization criterion, $\pi_0$ is better than $\pi_1$. However, when fixing a value $u$ less than both expected values, it can happen that $Pr\left(u^{\pi_0}(M) < u \big| \mathcal{B}\right) > Pr\left(u^{\pi_1}(M) < u \big| \mathcal{B}\right)$.

With this in mind, it can be useful to define risk-aware utility functions. Risk measures are widely studied in mathematical finance \cite{finance} since they are a way to rationally quantify risk. Their application to MDPs under model uncertainty was investigated in the work in \cite{risk}.

In the following, we introduce two risk measures: the Value at Risk $\left(\textit{VaR}\right)$ and the Conditional Value at Risk $\left(\textit{CVaR}\right)$ both inspired by the work in \citep{rockafellar2002conditional}.

Let $M$ be a random variable governed by a probability measure $Pr$ on its domain $\mathbb{M}$, and $u: \mathbb{M} \rightarrow \mathbb{R}$ be a measurable function such that $\mathbb{E}\big[ u(M) \big] < + \infty$. Following the work in \citep{rockafellar2002conditional}, we define the cumulative density function of $u(M)$ as:
\begin{equation}
\Psi(a) = Pr \big( u(M) \leq a \big)
\end{equation}
While in ﬁnance or insurance industry, losses that should be minimized (by looking for the optimal decision) are considered, in the MDP framework, the function $u$ is a utility function that should be maximized (by seeking the optimal strategy).
Let us consider a (low) risk level $q \in (0,1)$, that corresponds to the (high) confidence level in \citep{rockafellar2002conditional}.
\begin{definition}[Value at Risk]
\label{def:var}
The Value at Risk (\textit{VaR}) of the utility function $u$, at the risk level $q$ is
\begin{equation}
    a_{q} = \min \big\{a \in \mathbb{R} \vert \Psi(a)\geq q \big\}.
    \label{varq}
\end{equation}
\end{definition}
The minimum in \ref{varq} is attained because $\Psi$ is non-decreasing and right continuous.
The definition of the Conditional Value at Risk is slightly different from the one in \citep{rockafellar2002conditional}, because again, in the MDP context, the lowest gains (to be maximized) are considered, and not the highest losses (to be minimized) like in finance. 
\begin{definition}[Conditional Value at Risk]
\label{def:cvar}
Let the cumulative density function of $u(M)$ conditional to $\{u(M) \leqslant a_q \}$ be
\begin{equation}
\Psi_{q}(a) = \left\{\begin{array}{ccc}
    \dfrac{\Psi(a)}{\Psi(a_q)} & \text{ for } a \leq a_q,\\
    1 & \text{ for } a  > a_q.
    \end{array}\right.
\label{ccdf}
\end{equation}
The Conditional Value at Risk (\textit{CVaR}) of the utility function $u$ at risk level $q$ is the expectation of the variable drawn by the cumulative density function \ref{ccdf}
\begin{equation}
    \phi_{q} = \mathbb{E}_{X \sim \Psi_q} \left[X\right].
\end{equation}
\end{definition}
Note that $\Psi(a_q)$ may be higher than $q$, that is why $\Psi(a_q)$ is used instead of $q$ in the denominator of the Eq. \ref{ccdf}, following the rescaling solution proposed in \citep{rockafellar2002conditional} in case of a probability atom at $a_q$. With these definitions, the (Conditional) Value at Risk of the utility function $u$ at risk level $q$ is also the (Conditional) Value at Risk of $u$ at risk level $\Psi(a_q) \geq q$.

\begin{remark}
Since the space of possible transition functions, that can be sampled from a Dirichlet posterior $\tau_p$, has the cardinality of the continuum, the distribution of the performance with respect to the uncertainty is continuous.
\end{remark}

\subsection{Risk-aware Solutions to BMDPs}
In the following, the BMDP utility of \eref{eq:bayesian} is generalized to take the risk into account.

\begin{definition}%
\it{Let $\beta$ be a BMDP and let $V_M^{\pi}\left(s\right)$ be the value function at state $s$ while following a policy $\pi$ in the MDP $M$ with transitions distributed according to the posterior $\tau_p$. Let also $Pr\left(u^{\pi}(M)|\mathcal{B}\right)$ be the pdf over the possible values assumed by $u^{\pi}(M)=\mathbb{E}_{S\sim\mu_0}\left[V_M^{\pi}\left(S\right)\right]$ given the batch $\mathcal{B}$. Then a risk-aware utility function is defined as:}
\begin{equation}
    \label{eq:w}
    \mathcal{U}_{\beta,\sigma}^{\pi} \defeq \mathbb{\sigma}_{M\sim\tau_p}\left[u^{\pi}(M)\right]
\end{equation}
where, $\sigma$ is a risk measure.%
\end{definition}
\begin{remark}
As a consequence of Definition \ref{def:cvar} if $\sigma$ is the Conditional Value at Risk at risk level $1$ then the BMDP utility of \eref{eq:bayesian} is a particular case of \eref{eq:w}.
\end{remark}

\section{Solving Offline a Risk-aware BMDP}
\label{sec:evc}

The expectation over the distribution of models makes the solution of a BMDP an intractable computational task. Moreover, a Risk-aware BMDP also presents an additional difficulty: the risk measure requires an estimate of the quantiles of the unknown value distribution given a policy.
An analytical maximization of the performance defined in \eref{eq:w} is often either impossible or too computationally demanding. In order to tackle the maximization problem, a valuable choice is resorting to a Monte Carlo estimate of the performance. We will then look for a sub-optimal policy, rather than an optimal one, by constraining the search to a set of candidate policies $\Pi$.

However, what number $L_\pi \in \mathbb{N}$ of models would be necessary to be sampled in order to have an accurate estimate of the performance of a policy within a chosen confidence interval? Ideally, $L_\pi$ should be as small as possible because Policy Evaluation has to be performed $L_\pi$ times in order to obtain the Bayesian posterior distribution of values assumed by the Value Functions. 

Fortunately, this problem has been addressed by the work in \cite{quantile}. It proposes a procedure that allows iteratively sampling values from a distribution whose quantile is required until the estimate of the said quantile will fall within a confidence interval with a required probabilistic significance. 

\begin{figure}[!ht]
    \centering
    \begin{tikzpicture}[scale=0.9, transform shape]
        \draw [step=1.0,black!70, very thick] (1,1) grid (10,5);
        \foreach \i in {1,...,9} {
            \foreach \j in {1,...,4} {
            }
        }
        
        \node at (1.5,1.5) {$u_1$};
        \node at (1.5,2.5) {$u_1$};
        \node at (1.5,3.5) {$u_1$};
        \node at (1.5,4.5) {$u_1$};
        
        \node at (2.5,1.5) {$_{\dots}$};
        \node at (2.5,2.5) {$_{\dots}$};
        \node at (2.5,3.5) {$_{\dots}$};
        \node at (2.5,4.5) {$_{\dots}$};
        
        \node at (3.5,1.5) {$u_{r-1}$};
        \node at (3.5,2.5) {$u_{r-1}$};
        \node at (3.5,3.5) {$u_{r-1}$};
        \node at (3.5,4.5) {$u_{r-1}$};
        
        \node at (4.5,1.5) {$u_g$};
        \node at (4.5,2.5) {$u_g$};
        \node at (4.5,3.5) {$u_g$};
        \node at (4.5,4.5) {$u_g$};
        
        \node at (5.5,1.5) {$_{\dots}$};
        \node at (5.5,2.5) {$_{\dots}$};
        \node at (5.5,3.5) {$_{\dots}$};
        \node at (5.5,4.5) {$_{\dots}$};
        
        \node at (6.5,1.5) {$u_{h-1}$};
        \node at (6.5,2.5) {$u_{h-1}$};
        \node at (6.5,3.5) {$u_{h-1}$};
        \node at (6.5,4.5) {$u_{h-1}$};
        
        \node at (7.5,1.5) {$u_h$};
        \node at (7.5,2.5) {$u_h$};
        \node at (7.5,3.5) {$u_h$};
        \node at (7.5,4.5) {$u_h$};
        
        \node at (8.5,1.5) {$_{\dots}$};
        \node at (8.5,2.5) {$_{\dots}$};
        \node at (8.5,3.5) {$_{\dots}$};
        \node at (8.5,4.5) {$_{\dots}$};
        
        \node at (9.5,1.5) {$u_L$};
        \node at (9.5,2.5) {$u_L$};
        \node at (9.5,3.5) {$u_L$};
        \node at (9.5,4.5) {$u_L$};
        
        \node[draw, red!50!black] at (3.5,5.25) (r) {g};
        \node[draw, red!50!black] at (7.5,5.25) (s) {h};
        
        \draw[thick, blue] (2.5, 4.5) ellipse (1.45 and 0.45);
        \draw[thick, blue] (3, 3.5) ellipse (1.95 and 0.45);
        \draw[thick, blue] (3.5, 2.5) ellipse (2.45 and 0.45);
        \draw[thick,blue] (4, 1.5) ellipse (2.95 and 0.45);

        \draw[thick, green!50!black] (7, 4.5) ellipse (2.95 and 0.45);
        \draw[thick, green!50!black] (7.5, 3.5) ellipse (2.45 and 0.45);
        \draw[thick, green!50!black] (8, 2.5) ellipse (1.95 and 0.45);
        \draw[thick,green!50!black] (8.5, 1.5) ellipse (1.45 and 0.45);

        \node at (4,0.33) (rleft) {};
        \node at (7,0.33) (sright) {};
        \node at (10.88, 5) (uright) {};
        \node at (10.88, 1) (dright) {};
        \node at (10.88, 5) (uright) {};
        \draw[decorate,decoration=brace] (sright) -- (rleft);
        
        \draw[decorate,decoration=brace] (uright) -- (dright);

        \node at (12.58, 3) {\parbox{2cm}{Possible events}}; %
        
        \node at (5.5, 0) (text) {$u_g \leq a_{q} < u_h$};
        
        \draw[very thick, red!50!black] (4, 0.5) -- (4, 5.5);
        \draw[very thick, red!50!black] (7, 0.5) -- (7, 5.5);

    \end{tikzpicture}
    \caption{\label{fig:estim_expl}Example of how the estimate in Algorithm \ref{algo:nmcs} works: imagine you have an ordered list with $L$ values $u_i$, $i \in \{1,\dots,L\}$ represented in the figure as rows. The probability of the event in \eref{eq:rsprob} is $\binom{L}{g} q^g (1-q)^{L-g}$ and corresponds to the probability of the random variable $B$ defined in \eref{eq:bdef} to assume all integer values between $g$ and $h-1$. The said probability is the sum of the probability of the events $B=i$ with $ g \leq i < h$. In the figure, every addend is represented as a row. In blue are encircled the values of $u_i$ smaller than $a_q$ and in green the ones bigger. The algorithm looks for the indices $(g,h)=\textit{argmin}_{(g,h)} |g-h|$ such that $Pr(u_g \leq a_q < u_h) > 1-\alpha$ and $u_h - u_g < \varepsilon$ where $\varepsilon$ is an error term dictating the maximum acceptable size of the confidence interval.}
\end{figure}
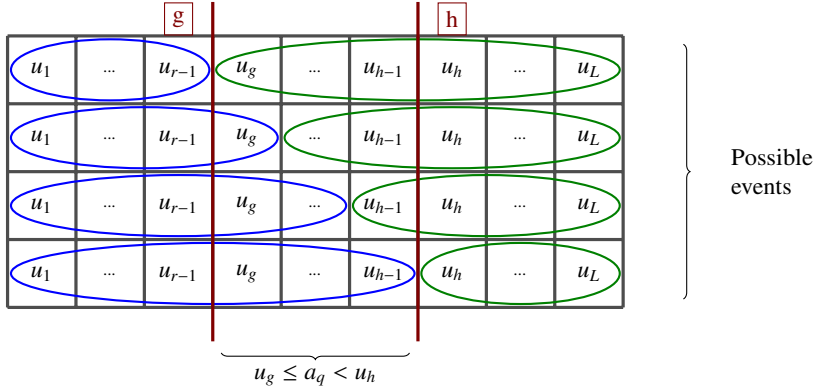

In the present work, we exploit the idea of estimating a quantile through sampling to propose the Monte Carlo Confident Policy Selection (MC2PS) algorithm. MC2PS is presented in Algorithm \ref{algo:nmcs}. MC2PS identifies a robust policy for a Risk-aware Bayesian MDP among a set of candidate policies.
\begin{algorithm}[htb!]
\caption{MC2PS}%
\label{algo:nmcs}
\hspace*{\algorithmicindent} \textbf{Input}: set of policies $\Pi$, significance level $\alpha \in \left[0,1\right]$, sampling batch size $k \in \mathbb{N}$, relative error tolerance $\varepsilon_{rel} \in (0,1]$, posterior distribution $\tau_p$, risk level $q \in \left(0,1\right)$, risk measure $\sigma$, initial state distribution $\mu_{0}$, evaluation discount factor $\gamma$.\\
\hspace*{\algorithmicindent}\textbf{Output:} best policy $\pi^*$.
\begin{algorithmic}[1]
\For{$\pi \in \Pi$}
\State $\mathcal{U}^{\pi}_{\beta,\sigma} \gets %
\textsc{RiskEvaluation}\left(\pi, \sigma, \tau_p, \mu_{0}, \varepsilon_{rel}, \alpha, q, k,\gamma\right)$ \EndFor
\State $ \textbf{return } \pi^* = \displaystyle \argmax_{\pi \in \Pi} \mathcal{U}^{\pi}_{\beta,\sigma}$
\\
\Procedure{RiskEvaluation}{}
\State \textbf{Input}: policy $\pi$, risk measure $\sigma \in \{ VaR, CVaR \}$, posterior distribution $\tau_p$, initial state distribution $\mu_{0}$, relative error threshold $\varepsilon_{rel} \in \left[0,1\right]$, significance level $\alpha \in \left[0,1\right]$, risk level $q \in \left(0,1\right)$, sampling batch size $k \in \mathbb{N}$, evaluation discount factor $\gamma$.
\State Initialize $u^{\pi} = \varnothing$ %
\State \textit{(the %
loop estimates the quantile needed in \eref{eq:w}%
)}
\Repeat
\For{$j\in \set{1, \ldots, k}$ \textbf{in parallel}}
\State Sample $\mathcal{M}_j \sim\tau_p$
\State $V^{\pi}_{\mathcal{M}_{j}}\left(s\right) \gets$ Policy Evaluation on model $\mathcal{M}_j$
\State $u^{\pi}(\mathcal{M}_{j}) \gets \mathbb{E}_{S\sim\mu_0}[ V^{\pi}_{\mathcal{M}_j}(S)]$ \eref{eq:performance}
\State $u^{\pi} \gets$ \textbf{ append } $u^{\pi}(\mathcal{M}_{j})$
\EndFor
\State $L_{\pi} \gets \lvert u^\pi \rvert$ %
\State Sort $u^{\pi}$ in increasing order
\State Find %
$\left(g,h\right) \in \mathbb{N}^{2}$ such that $|h-g|$ is minimal and:
\State \hspace{1cm} $Pr(u^\pi_g \leq a_q < u^\pi_h) = \left(\sum_{i=g}^{h-1}\binom{L_{\pi}} {i}q^{i}\left(1-q\right)^{L_{\pi}-i}\right) > 1-\alpha$;
\Until{:}
\State \hspace{1cm}$u^{\pi}_{h} - u^{\pi}_{g} < \varepsilon_{rel} \cdot \left(u^{\pi}_{L_{\pi}}-u^{\pi}_{1}\right)$%
\State \textbf{if} $\sigma = VaR$ \textbf{then}
\State \hspace{1cm} $\widehat{a_q} \gets u^{\pi}_{g}$
\State \hspace{1cm}\textbf{return} $\widehat{a_q}$
\State \textbf{if} $\sigma = CVaR$ \textbf{ then}
\State \hspace{1cm} $\widehat{\phi_q} \gets \frac{1}{g}\sum_{i=1}^{g}u^{\pi}_{i}$
\State \hspace{1cm}\textbf{return} $\widehat{\phi_q}$ %
\EndProcedure
\end{algorithmic}
\end{algorithm}
In detail, for a given set of policies $\Pi$ and for every policy $\pi \in \Pi$, the algorithm incrementally samples $k$ transition models from $\tau_p$ and performs Policy Evaluation in parallel for each one of them until the stopping criterion is reached (see the \textsc{RiskEvaluation} procedure in Alg. \ref{algo:nmcs}). %
The stopping criterion guarantees that the estimate of the $q$-quantile is statistically well approximated with a significance level $\alpha$ within a dynamically sampled confidence interval whose width is smaller than $\varepsilon_{rel}$ (lines 19-22) given the total $L_\pi$ models sampled.\\
Indeed, being $u^{\pi}_{i} := u^{\pi}(M_i)$ the list of ordered performance values obtained from the sampled $L_\pi$ models with $i \in \{1,\dots,L_{\pi}\}$, the probability that the elements with indices $h$ and $g$ of this list are bounding $a_q$, will be given by the probability of the union of all the (incompatible) events that lead to $u_{g}^{\pi}\leq a_q \leq u_{h}^{\pi}$ (see Figure \ref{fig:estim_expl}).
In detail, let $a_q$ be the theoretical Value at Risk of $u^{\pi}(M)$ at risk level $q$. Let us denote sampled utility values in increasing order by $u^{\pi}(M_1) \leq \ldots \leq u^{\pi}(M_{L_{\pi}})$, and suppose that the utility distribution has no probability atom at $a_q$: $\forall 1 \leq i \leq L_{\pi}$, $\Psi(a_q) := Pr\big( u^{\pi}(M_i) \leq a_q \big) = q$. 
    Let us introduce the random variables
\begin{equation}
\label{eq:bdef}
B_i = \mathds{1}_{\big\{ u^{\pi}(M_i) \leq a_q \big\}} = 
\begin{cases}
    1, & \text{if } u^{\pi}(M_i) \leq a_q, \\
    0, & \text{otherwise.}
\end{cases}
\end{equation}
The random variables $B_i$ are drawn by a Bernoulli distribution with parameter $q$. The random variable $B = \sum_{i=1}^{L_{\pi}}B_i$ is the number of sampled utilities that are lower than $a_q$, drawn by a binomial distribution with parameters $L_{\pi}$ and $q$. The event $\big\{ u^{\pi}(M_g) \leq a_q < u^{\pi}(M_h) \big\}$ is $\cup_{i=g}^{h-1}\{ B = i\}$, \textit{i.e.} the event ``there are exactly $g$, $g+1$, \ldots, or $h-1$ sampled utility values that are lower than $a_q$''. Using the binomial distribution formula, the probability of this event is
\begin{equation} Pr \Big( u^{\pi}(M_g) \leq a_q < u^{\pi}(M_h) \Big) =
    \sum_{i=g}^{h-1} Pr ( B = i )  = \sum_{i=g}^{h-1} \binom{L_{\pi}}{i} q^i (1-q)^{L_{\pi}-i}.\end{equation}
Hence by imposing constraint $\sum_{i=g}^{h-1} \binom{L_{\pi}}{i} q^i (1-q)^{L_{\pi}-i} > 1-\alpha$ when selecting indices $r$ and $s$, we ensure that
\begin{equation}
\label{eq:rsprob}
 Pr \Big( u^{\pi}(M_g) \leq a_q < u^{\pi}(M_h) \Big)>1-\alpha,   
\end{equation}
\textit{i.e.} we get probabilistic bounds computed from the sampled utility values.\\
Note that, if there is a probability atom at $a_q$, \textit{i.e.} $Pr\big(u^{\pi}(M_i) = a_q\big) > 0$, the previous reasoning cannot be applied directly in the case where $q < \Psi(a_q) := {Pr\big(u^{\pi}(M_i) \leq a_q \big)}$. However, we can write
\begin{equation}q_{-} := Pr \big( u^{\pi}(M_i) < a_q \big) \leq q <  q_{+} := Pr \big( u^{\pi}(M_i) \leq a_q \big),\end{equation}
and one can show that selected indices $g$ and $h$ are non decreasing with $q$. Thus, using the risk level $q$, selected indices are higher than those that would be selected using $q_{-}$, and lower than those that would be selected using $q_+$, both corresponding to utility values bounding the location of the probability atom, \textit{i.e.} the Value at Risk $a_q$, with probability $1-\alpha$.
    
Eventually, the algorithm leverages the estimate of both the Value at Risk and of the policy value achieved on sampled models to obtain an estimate of the utility function $\mathcal{U}_{\beta, \sigma}^{\pi}$ for a specific risk measure $\sigma$ and risk level $q$. For instance, return the estimate of  $a_q$ if $\sigma = VaR$ or $\phi_q$ if $\sigma = CVaR$ (lines 23-28).
Finally, once the utility function has been estimated for every policy, it outputs the one that maximizes it (line 4).

\begin{remark}
Let $\Lambda$ be the total number of models sampled to estimate the quantile of the performance distribution among policies: $\Lambda = \sum_{\pi\in\Pi} L_{\pi}$.
MC2PS performs Policy Evaluation $\Lambda$ times. The size of the space of all applicable policies of a finite state and action space MDP is $\lvert\Pi\rvert = \lvert \mathcal{A} \rvert ^{\lvert \mathcal{S} \rvert}$. It goes without saying that looking over the whole policy space can be practically intractable even for not-so-big MDPs. Nevertheless, restricting the research to a subset of policies could be a viable solution also for big MDPs, also considering that the Policy Evaluations are carried out in parallel.
\end{remark}
\subsection{Exploitation vs Caution (EvC)}
Reference \cite{gamma} shows that the policy obtained by solving an MDP $\hat{M}=(\mathcal{S},\mathcal{A},\hat{T},r,\gamma^*)$, trivially learned from a batch of experiences collected from another MDP $M=\left(\mathcal{S},\mathcal{A},T,r,\gamma_{ev}\right)$, with $\gamma^*$ a discount factor such as $\gamma^*\leq \gamma_{ev}$, is more efficient in $M$ than the policy obtained by solving $\hat{M}$ using $\gamma_{ev}$. The reason is that $\hat{T}$ is an approximation of $T$ and may not be trusted for long-term planning horizons.
The selection of the best $\gamma^*$ optimizes a trade-off between the exploitation of the information contained in the batch and the necessity of being cautious since the model estimate is not perfect.

Inspired by the conclusions of reference \cite{gamma}, and also guided by the intuition that the model $M$ that generated the data will be different from $\hat{M}$, but hopefully \textit{close} to it, we expect that the policies obtained by solving another MDP $\tilde{M} =  (\mathcal{S},\mathcal{A},\tilde{T},r,\gamma)$ with $\tilde{T}$ \textit{close} to $\hat{T}$ and $\gamma \leq \gamma_{ev}$ can be viable solutions for the Risk-aware Bayesian MDP.

Henceforth, the Exploitation \textit{versus} Caution (EvC) algorithm is presented and schematized in Algorithm \ref{algo:evc}. %
EvC will search for a promising risk-aware policy by focusing the search on a set of candidate policies $\Pi$ computed with several baseline algorithms $\mathbb{A}$, which is further enriched by solving different MDPs $\tilde{M}$ and $\gamma\leq\gamma_{ev}$ values. Remember that the goal is to find a policy that is performant in the model $M$ taking into account that the agent does not have access to the probability values that define the model, but only to the pre-collected batch. Model uncertainty is framed within a Bayesian MDP. As already stated, we do not aim to find the optimal solution to the BMDP, but rather to select the best policy in terms of robustness within the ones in the candidate set.

\begin{algorithm}[!t]
\caption{EvC} %
\label{algo:evc}
\hspace*{\algorithmicindent} \textbf{Input}: risk level $q \in \left[0,1\right]$, significance level $\alpha \in \left[0,1\right]$, sampling batch size $k \in \mathbb{N}$, relative error tolerance $\varepsilon_{rel} \in \left[0,1\right]$, posterior distribution $\tau_p$, risk measure $\sigma$, initial state distribution $\mu_{0}$, set of discount factors $G$, number of models to solve $l \in \mathbb{N}$, $\mathcal{B}$ batch of transitions, evaluation discount factor $\gamma_{ev}$\\
\hspace*{\algorithmicindent}\textbf{Output:} best policy $\pi^*$.
\begin{algorithmic}[1]
\State $\Pi \gets \textsc{GeneratePolicies}\left(\tau_p, G, l\right)$
\State \textbf{return} $\pi^* = \text{MC2PS}\left(q, \alpha, k, \varepsilon_{rel}, \tau_p, \Pi, \sigma, \mu_{0}, \gamma_{ev}\right)$ \label{alg:launchMC2PS}%
\\
\Procedure{GeneratePolicies}{}
\State \textbf{Input}: posterior distribution $\tau_p$, set of discount factors $G$, number of models $l \in \mathbb{N}$ to be solved, $\mathcal{B}$ batch of transitions.
\State Initialize $\mathbb{M} = \left\{l \text{ transition models} \sim \tau_p\right\} \cup \{\hat{T}\}$ \label{alg:transmodels} %
\State Initialize $\Pi = \varnothing$ (an empty set)
\State Initialize $\mathbb{A} = \{\textit{SPIBB}, \textit{BOPAH}, \textit{BCR}, \textit{NORBU}\}$ (examples of baseline algorithms) \label{alg:portfolio}
\For{$\left(\gamma \in G, T \in \mathbb{M}\right)$} \label{alg:gamma1}
\State $\pi_{\left(T,\gamma\right)} =$ solution to the MDP with $T$ and $\gamma$
\State Append $\pi_{\left(T,\gamma\right)}$ to $\Pi$ if 
$\pi_{\left(T,\gamma\right)} \not\in \Pi$ \label{alg:append1}
\EndFor \label{alg:gamma2}
\For{$\textit{algorithm} \in \mathbb{A}$}
\State $\pi_{algorithm} =$ solution to the Offline MDP with $\mathcal{B}$ and $\textit{algorithm}$
\State Append $\pi_{\textit{algorithm}}$ to $\Pi$ if $\pi_{algorithm} \not\in \Pi$ \label{alg:append2}
\EndFor
\State \textbf{return} $\Pi$
\EndProcedure
\end{algorithmic}
\end{algorithm}
In detail, EvC first generates candidate policies that will constitute the set $\Pi$ (line 1 calls \textsc{GeneratePolicies} procedure). %
For this, starting from the batch the problem is solved using a portfolio constituted of state-of-the-art algorithms (line \ref{alg:portfolio}). On top of that the trivial MDP\footnote{The trivial MDP $\hat{M}$ is a straightforward MDP estimate using the batch $\mathcal{B}$. For instance, in the case of a discrete MDP this is equivalent to the model that maximizes the likelihood of $\mathcal{B}$, \textit{i.e.} the one whose transition probabilities are obtained from the frequencies of transitions in the batch.} $\hat{M}$ and $l$ additional MDPs are sampled from the Bayesian posterior $\tau_p$  obtained from the batch (line \ref{alg:transmodels}), and then solved with different values of $\left\{\gamma \in G \vert \gamma \leq \gamma_{ev}\right\}$ (lines \ref{alg:gamma1}-\ref{alg:gamma2}). Recalling that $\gamma_{ev}$ is the discount factor of the Risk-aware Bayesian MDP. Note that the obtained set $\Pi$ has unrepeated solutions (line \ref{alg:append1} and line \ref{alg:append2}). As a last step, MC2PS is launched with the obtained set of candidate policies $\Pi$ returning the best risk-aware solution $\pi^* \in \Pi$ (line \ref{alg:launchMC2PS}). 

\begin{remark}
Note, if we test over $9$ different discount factors, such as $G = \left\{0.1, 0.2, \dots, \gamma_{ev} = 0.9\right\}$, and 5 different $(l = 5)$ MDPs $\tilde{M}$ (including $\hat{M}$), then we solve $\lvert\Pi\rvert\leq 9l = 45$ MDPs to enrich the set of candidate policies within this approach. 
\end{remark}

\subsection{Theoretical guarantees}
\label{sec:theo}
Since EvC searches for the policy $\pi \in \Pi$ that maximizes the criterion of Eq. \eqref{eq:w}, the Algorithm \ref{algo:evc}, rather than yielding a sub-optimal solution to the Risk-aware BMDP, can be seen as a policy selection approach. Assuming that the Bayesian posterior $\tau_p$ efficiently encodes the model uncertainty, EvC outputs a policy whose performance in the real environment is guaranteed in probability to be greater than some value that changes with respect to the chosen risk-aware measure. In a simpler way, we can provide theoretical guarantees on the estimate of the quantile needed to compute the risk-aware utility function that will be eventually maximized over the set of candidate policies.

\begin{theorem}
\label{garant}
Let $\pi \in \Pi$ be a candidate policy and $u^{\pi}(M_g)$ be an estimate of the Value at Risk of $u^{\pi}(M)$ at risk level $q$ calculated through EvC. Let $u^{\pi}(M)$ be the performance of $\pi$ with $M$ distributed according to the Bayesian posterior $\tau_p$.
The performance of $\pi$ in this MDP $M$ 
is greater than the estimate of $a_q$ with probability: 
\begin{equation}
    \label{eq:thm1}
    Pr\big(u^{\pi}(M) \geq u^{\pi}(M_g) \big) \geq (1-q)(1-\alpha). 
\end{equation}
\begin{prf}
Note that $\big\{u^{\pi}(M) \geq a_q\big\} 
\cap 
\big\{a_q \geq u^{\pi}(M_g)\big\}
\subseteq 
\big\{ u^{\pi}(M) \geq u^{\pi}(M_g)\big\}$, where $a_q$ denotes the Value at Risk of $u^{\pi}(M)$ at risk level $q$. The two events of the intersection respectively depend on two independent random variables -- a future performance $u^{\pi}(M)$, that could be obtained by acting according to the policy $\pi$, and a Value at Risk estimate $u^{\pi}(M_g)$, whose randomness is the result of the sampling procedure in the Algorithm \ref{algo:nmcs}.
The previous %
inclusion allows writing
$Pr\big(u^{\pi}(M) \geq u^{\pi}(M_g)\big) \geq Pr\big( u^{\pi}(M) \geq a_q \big)
\cdot
Pr\big(a_q \geq u^{\pi}(M_{r})\big)
\geq
(1-q) \cdot Pr\big(a_q \geq u^{\pi}(M_g)\big)
\geq (1-q) (1-\alpha) $.
The last inequality is ensured by the quantile estimation (lines 19-22 in Algorithm 1), and the previous one by the definition of $a_q$.
Therefore, we get the equation \ref{eq:thm1}.
\end{prf}\end{theorem}

\begin{remark}
When the risk-aware measure used in EvC is $VaR$ the lower bound on $u^{\pi}(M)$ ($u^{\pi}(M_g)$) in the Proof of Theorem 1 is maximized over the policies. If the risk-aware measure is $CVaR$ the empirical expected value over the $q$-fraction of low-performant policies is maximized.
\end{remark}
\begin{remark}
Since $u^{\pi}(M_1) \leq u^{\pi}(M_2) \leq \ldots \leq u^{\pi}(M_g)$, then $\frac{1}{g}\sum_{i=1}^{g}u^{\pi}(M_i) \leq u^{\pi}(M_g)$,
therefore the same lower bound in probability is also valid for the \textit{CVaR} utility function:
\begin{equation}
    Pr\left(u^{\pi}(M) \geq \frac{1}{g}\sum_{i=1}^{g}u^{\pi}(M_i) \right) \geq Pr\big(u^{\pi}(M) \geq u^{\pi}(M_g) \big) \geq (1-q)(1-\alpha). 
\end{equation}
\end{remark}

Note that the sampling procedure in the Algorithm \ref{algo:nmcs} ensures that $Pr\big( \lvert u^{\pi}(M_g) - a_q \rvert \geq \varepsilon \big) \leq \alpha$, with $\varepsilon = \varepsilon_{rel} \cdot (u^{\pi}(M_{L_{\pi}})-u^{\pi}(M_1))$. If a practitioner wants such a probabilistic bound on the precision of the estimate of $\phi_q$, she/he should sample additional models $N \in \mathbb{M}$ from the posterior, to select $n$ independent models such that $u^{\pi}(N) \leq u^{\pi}(M_g)$, %
where $u^{\pi}(M_g)$ is given by the sampling procedure of Algorithm \ref{algo:nmcs}.
The new estimate of $\phi_q$ computed from these $n$ new models benefits from the following theorem.
\begin{theorem}
\label{garant2}
Let $\pi \in \Pi$ be a candidate policy, $N_i \in \mathbb{M}$ be one of the $n$ new sampled models from the posterior $\tau_p$ such that $\forall i, u^{\pi}(N_i) \leq u^{\pi}(M_g)$, with $u^{\pi}(M_g)$ calculated through EvC, and $\overline{U} = \frac{1}{n}\sum_{i=1}^{n}u^{\pi}(N_i)$ be the new estimate of $\phi_{q}$. This new estimate of the Conditional Value at Risk of $u^{\pi}$ at risk level $q$ respects the following inequality:%
\begin{equation}
    Pr\left(\lvert \overline{U} - \phi_q \rvert \geq t \right) 
\leq 2 \exp \left( - \dfrac{2nt^2}{ \big( u^{\pi}(M_g) - \xi\big)^2}\right) + \alpha,
\end{equation}
with $\xi = \inf_{m \in \mathbb{M}}u^{\pi}(m)$, or any other lower bound of $u^{\pi}$ as, for instance, $\frac{r_{min}}{1-\gamma}$. Note that $\xi\geq 0$ if the reward values are known to be non-negative.

\end{theorem}
\begin{prf}
By using the law of total probability,
and upper bounding some probability values by 1,
\begin{eqnarray*}
Pr\Big(\lvert \overline{U} - \phi_q \rvert \geq t \Big) & = & Pr\Big( \lvert \overline{U} - \phi_q \rvert \geq t \big\vert \forall i, u^{\pi}(N_i) \leq a_q \Big) Pr(\forall i, u^{\pi}(N_i) \leq a_q)\\
& & \hspace{0.2cm} +  \hspace{0.2cm}  Pr\Big( \lvert \overline{U} - \phi_q \rvert \geq t \big\vert \exists i \mbox{ s.t. } u^{\pi}(N_i) > a_q \Big)
Pr\big( \exists i \mbox{ s.t. }u^{\pi}(N_i) > a_q\big)\\
& \leq & Pr\Big( \lvert \overline{U} - \phi_q \rvert \geq t \big\vert \forall i, u^{\pi}(N_i) \leq a_q \Big) + Pr\big( \exists i \mbox{ s.t. }u^{\pi}(N_i) > a_q\big).
\end{eqnarray*}
The probability value on the right is lower than $Pr\big( u^{\pi}(M_g) > a_q \big) \leq Pr\Big(a_q \notin \big[u^{\pi}(M_g),u^{\pi}(M_h)\big)\Big) \leq \alpha $ using the inequality $Pr (u^{\pi}(M_g) \leq a_q) > 1 - \alpha$ from lines 19-22 of Algorithm \ref{algo:nmcs}.
What follows only depends on the definition of $\phi_q$ as the expected value up to $a_q$, and Hoeffding's inequality.

\end{prf}
\subsection{Consequences and applications}
\label{sec:consequences}
The purpose of Offline Learning is that of %
providing behavioral policies to be applied by real-world automated agents%
. Thus reducing the risk at the expense of a longer computational phase is not only commendable but compulsory. Will the policy obtained through MC2PS and EvC be \textit{good} or entirely-risk free? This goes beyond the theoretical guarantees provided by the algorithms since its outputs depend not only on the characteristics of the environment and on the set of candidate policies but also on the quality and variety of the batch. A batch of transitions that is too small or too concentrated in the same region of the state-action space may result in policies that, even if they are guaranteed to handle the risk better than the trivial one, can still be catastrophic. %
\section{Experiments}
\label{sec:exp}
In order to evaluate the proposed approach, we selected three small MDPs and hence easy-to-study stochastic environments endowed with diversified characteristics: two planning environments without absorbing states, Ring (5 states, 3 actions), and Chain (5 states, 2 actions). The former consisting in the stabilization of the agent in a particular non-absorbing goal with stochastic drift and the latter presenting cycles; and the Random Frozen Lake (RFL) environment, a re-adaptation of Frozen Lake from Open AI Gym suite \cite{gym} ($8 \times 8$ grid world with fatal absorbing states).

\subsection{Environments' description}
\label{app:env}

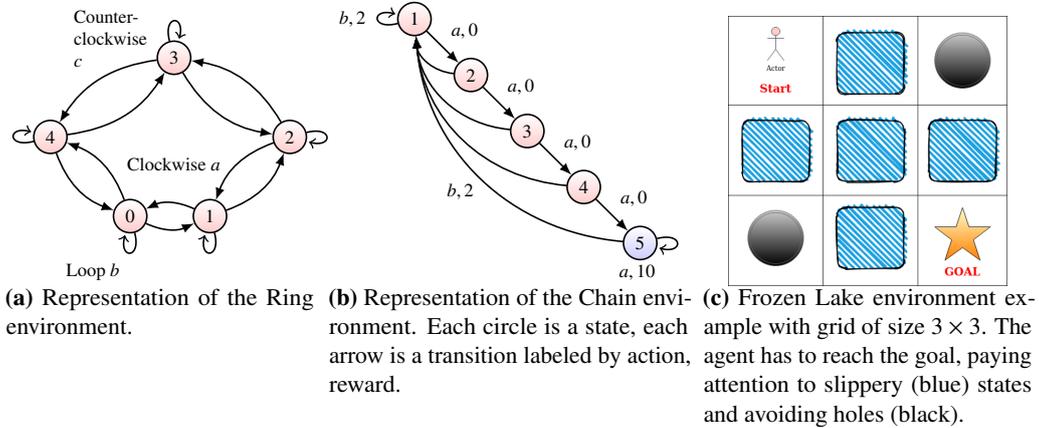
\begin{figure}
     \centering
     \begin{subfigure}[t]{0.3\textwidth}
         \centering
         \begin {tikzpicture}[scale=0.7, transform shape, -latex, auto, node distance =1.5 cm and 1.5 cm, on grid, semithick,
state/.style ={circle, top color = white, bottom color=red!20,
draw, black, text=black, minimum width = 0.5 cm},
goal/.style ={circle, top color = white, bottom color=blue!20,
draw, black, text=black, minimum width = 0.5 cm}]

\node[state] (0) {0};
\node[state] (1) [right=of 0] {1};
\node[state] (2) [above right=of 1] {2};
\node[state] (4) [above left=of 0] {4};
\node[state] (3) [above of=4, right=2cm] {3};

\path (1) edge [bend right =25] (2);
\path (2) edge [bend right =25] node[left=0.2cm] {Clockwise $a$} (1);

\path (2) edge [bend right =25] (3);
\path (3) edge [bend right =25] (2);

\path (3) edge [bend right =25] node[left=0.2cm, above=0.1cm] {\parbox{1cm}{Counter-clockwise $c$}} (4);
\path (4) edge [bend right =25] (3);

\path (4) edge [bend right =25] (0);
\path (0) edge [bend right =25] (4);

\path (0) edge [bend right =25] (1);
\path (1) edge [bend right =25] (0);

\path (0) edge [loop below] node[below left=0.1cm] {Loop $b$} (0);
\path (1) edge [loop below] (1);
\path (2) edge [loop right] (2);
\path (3) edge [loop above] (3);
\path (4) edge [loop left] (4);

\end{tikzpicture}
\caption{Representation of the Ring environment.}
\label{fig:ring-env}
         \label{fig:ring}
     \end{subfigure}
     \hfill
     \begin{subfigure}[t]{0.35\textwidth}
         \centering
         \begin {tikzpicture}[scale=0.7, transform shape, -latex, auto, node distance =1.5 cm and 1.5 cm, on grid, semithick,
state/.style ={circle, top color = white, bottom color=red!20,
draw, black, text=black, minimum width = 0.5 cm},
goal/.style ={circle, top color = white, bottom color=blue!20,
draw, black, text=black, minimum width = 0.5 cm}]

\node[state] (1) {1};
\node[state] (2) [below right=1.5cm of 1] {2};
\node[state] (3) [below right=1.5cm of 2] {3};
\node[state] (4) [below right=1.5cm of 3] {4};
\node[goal] (5) [below right=1.5cm of 4] {5};

\path (1) edge node[above right = 0.05cm] {$a,0$} (2);
\path (2) edge node[above right = 0.05cm] {$a,0$} (3);
\path (3) edge node[above right = 0.05cm] {$a,0$} (4);
\path (4) edge node[above right = 0.05cm] {$a,0$} (5);

\path (1) edge [loop left] node[left =0.1 cm] {$b,2$} (1);
\path (5) edge [loop right] node[below left =0.4 cm] {$a,10$} (5);

\path (2) edge [bend left =40] (1);
\path (3) edge [bend left =40] (1);
\path (4) edge [bend left =40] (1);
\path (5) edge [bend left =40] node[] {$b,2$} (1);
\end{tikzpicture}
\caption{Representation of the Chain environment. Each circle is a state, each arrow is a transition labeled by action, reward.}
\label{fig:chain-env}
         \label{fig:chain}
     \end{subfigure}
     \hfill
     \begin{subfigure}[t]{0.32\textwidth}
         \centering
         \includegraphics[width=0.85\linewidth]{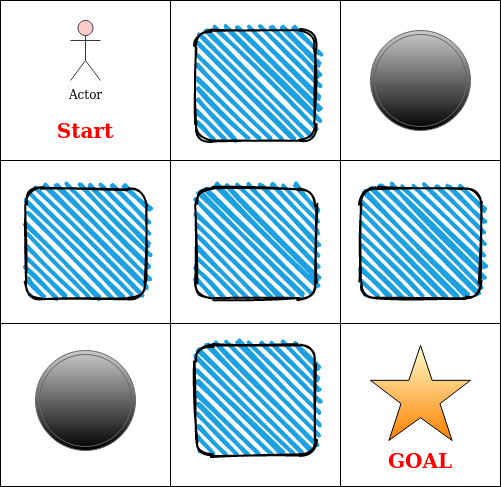}
         \caption{Frozen Lake environment example with grid of size $3\times3$. The agent has to reach the goal, paying attention to slippery (blue) states and avoiding holes (black).}
         \label{fig:frozen}
     \end{subfigure}
        \caption{Environments illustration.}
        \label{fig:envs}
\end{figure}

\paragraph{Ring} %
This environment is described by five states: $\{0,\dots,4\}$, forming a single loop. Three actions are possible: \textit{a}, \textit{b}, and \textit{c}. The agent starts in state $0$. The action \textit{a} will move it to the state $s-1$ with probability 1.0 (e.g. when in 4 it moves to 3) if $s=0,1,3$,  and with probability $0.5$ if it is elsewhere. With the action \textit{b} the agent will remain in the same state with probability $0.8$ and move to the left or to the right with probability $0.1$ if it is in state $s=0,1,3$, if it is in state $2$ or $4$ it will move with probability $1$. The action \textit{c} will move the agent to the right with probability $0.9$ and it will not move with probability $0.1$ if it is in state $s=0,1,3$. Otherwise, the same effects will apply, but with probability $0.5$. The agent earns an immediate reward $r=0.5$ if it moves from $2\rightarrow3$ or $4\rightarrow3$, and $r=1$ for any transition $3 \rightarrow 3$. Elsewhere $r=0$. A graphical representation is shown in Figure \ref{fig:ring-env}.

\paragraph{Chain} This environment was proposed in the research in \cite{strens} and was adapted to the present study. There are five states with the topology of an open chain and two actions $a$ and $b$. The agent starts from the state most to the left. With action $a$ the agent moves to the right and receives an immediate reward $r=0$ with probability $0.8$. Once the agent is in the rightmost state, performing the first action lets him stay there and receive a reward $r=10$ with probability $0.8$. It slips back to the origin earning a reward $r=2$ with probability $0.2$. Action $b$ moves the agent to the origin state with probability $0.8$ receiving a reward $r=2$ or letting it go right with probability $0.2$ earning $r=0$. The optimal policy consists of applying action $b$ in the first state and action $a$ in the others.  A representation is shown in Figure \ref{fig:chain-env}.

\paragraph{Random Frozen Lake (RFL)} The Frozen Lake Environment of the Open AI Gym suite \cite{gym} was edited for this study. The agent moves in a grid world ($8\times8$). It starts in the utmost left corner and it must reach a distant absorbing goal state that yields a reward $r=1$. In the grid there are some holes. If it falls into a hole it is blocked there and it can not move anymore, obtaining from that moment an immediate reward $r=0$. Unfortunately, the field is covered with ice and hence it is slippery. When the agent wants to move towards a nearby state it can slip with fixed probability $p$ and ends up in an unintended place. The grid is generated randomly assuring that there always exists a hole-free path connecting the start and the goal. Moreover, to each couple of action and non-terminal state $\left(a, s\right)$ is assigned a different immediate reward $r$ sampled at random between $\left(0, 0.8\right)$ at the moment of the generation of the MDP problem. The MDP itself does not have a stochastic reward, but the map and the rewards are randomly generated. A graphical representation (for a $3\times 3$ grid) is shown in Figure \ref{fig:frozen}.

\subsection{Setup}
Given $\left(n, m\right) \in \mathbf{N}^{2}$, $m$ trajectories, with $n$ steps each, are generated following a random policy in each environment. We opted for a random data collecting procedure because we imagine using EvC in a scenario where both the developers and the autonomous agent are completely agnostic about the model dynamics and have no prior knowledge.

The true environment is assumed to be known for the a posteriori evaluation. The most likely transition model is inferred from the batch. The trivial MDP was then solved with the Policy Iteration algorithm and its relative performance in the true environment is obtained by Policy Evaluation. EvC data was computed with $\phi_{0.25}$ and $a_{0.25}$ (the first quartile). For each of these risk-aware measures, the following parameters (see Algorithm \ref{algo:evc}) were used: the set of discount factors $G=\left\{0.2, 0.4, 0.6, 0.8, 0.9\right\}$, the significance level $\alpha = 0.01$, the relative tolerance error $\varepsilon_{rel}=0.01$, and the number $l$ of different models sampled from the prior is given in Table \ref{exp_params}.

In the experiments, for a given batch size $N = nm \in \mathbf{N}$, $50$ different batches were generated containing fixed size trajectories. The trajectory sizes used are also given in Table \ref{exp_params}.

The chosen state-of-the-art algorithms that provide the base for the set of candidate policies are the following:
\begin{enumerate}
    \item \textit{Deterministic policies:} output by the following baselines\footnote[1]{The code was taken from the authors' Github repository: \url{https://github.com/marekpetrik/craam2/tree/master/examples/evaluation/algorithms} and readapted.}, please notice that the quantile used for the robust and soft robust objectives in the algorithms is the same provided as general input for the estimate of EvC: \textbf{BCR} \cite{petrik2019beyond}, \textbf{NORBU - Soft Robust \textit{CVaR}} \cite{lobo2021softrobust} (soft robust hyperparameter $\lambda = 0.5$);
    \item \textit{Stochastic policies:} output by the following algorithms \footnote[2]{The code was taken from the Github repository: \url{https://github.com/KAIST-AILab/BOPAH} and readapted.}: \textbf{SPIBB} \cite{spibb} receiving as input the batch collector policy, and, \textbf{BOPAH} \cite{bopah} receiving as input the batch collector policy.
\end{enumerate}
In our implementation of these baselines we
only used intuitively tunable parameters (e.g. the discount factor).

\begin{remark}
We did not use MOPO \cite{MOPO} and MOReL \cite{MOReL} since:
(1) they have usually been tested on continuous state MDPs driven by deterministic dynamics, while here we are tackling non-deterministic environments; (2) they highly rely on hyperparameter domain-dependent fine tuning which we did not do to fulfill the offline learning obligation.
\end{remark}

In the evaluation phase, the discount factor is defined as $\gamma_{\text{ev}} = 0.9$. The others simulation parameters are provided in Table \ref{exp_params}. Eventually, we also compared EvC with UnO by performing the risk-sensitive off-policy evaluation with UnO over the same set of candidate policies provided to EvC and then selected the one that maximized the risk-sensitive objective. While it is true that UnO, as other Importance Sampling based off-policy evaluation methods, should not be able to accurately evaluate deterministic policy, we still compare our approach to it because there are no other risk-sensitive off-policy evaluation approaches of our knowledge.

\begin{table*}[t!]
\caption{Parameters and hyperparameters used during the simulations: $n$ is the number of steps in each trajectory contained in a batch; $l$ is the number of different models sampled from the prior in EvC (Algorithm \ref{algo:evc}); $\textbf{$\{N_{\wedge}\}$}$ is the set of different thresholds used in SPIBB; $\textbf{fold}$ and $\textbf{DOF}$ are the fold and degree of freedom hyper-parameters used in BOPAH; $\lambda$ is the soft robust hyper-parameter of NORBU. Bold values are displayed in the plots.}
\centering
\small
\begin{tabular}{lcccccc}
\hline
\textbf{Environment} & $n$ & $l$ &
  \textbf{$\{N_{\wedge}\}$} &  \textbf{fold} & \textbf{DOF} & \textbf{$\lambda$}\\ \hline
Ring  & $8$ & $3$& $\left\{\textbf{1}, 2, 3, 5, 7, 10, 20\right\}$ & $2$ & $20$ & $0.5$\\ \hline
Chain & $8$ & $3$ & $\left\{\textbf{1}, 2, 3, 5, 7, 10, 20\right\}$ & $2$ & $20$& $0.5$\\ \hline
RFL $\left(8\times8\right)$& $15$ & $10$ &  $\left\{\textbf{1}, 2, 3, 5, 7, 10, 20\right\}$ & $2$ & $20$ & $0.5$\\ \hline
\end{tabular}
\label{exp_params}
\end{table*}

\subsection{Metrics}
We report metrics about the performance differences $\Delta U = u^{\pi}_{\beta,\sigma} - u^{\pi_{trivial}}_{\beta,\sigma}$ of the policies $\pi$ obtained with a specific algorithm (\eref{eq:performance} using the utility function defined in \eref{eq:w}) and the performance obtained by solving the trivial model in the same setting and using the same batch of trajectories. This last value is normalized by the performance of the optimal policy.
In particular, we consider: (1) the maximal $\Delta U$ obtained, (2) the mean value over all the different simulations, (3) the median over all simulations, and (4) the minimal $\Delta U$.
The selected metrics provide insight into the validity of the approaches. We consider only the extrema of the distributions of the results (min, max), their median, and mean values since trying to estimate the whole distributions, and hence their quantiles could result in wrong conclusions if we are not sampling enough batches.
For instance, in order to correctly estimate the Value at Risk at risk level $q=0.25$ with a $\alpha = 0.01$ significance usually tens of thousands of samples are required.
However, we are performing only hundreds of simulations with a fixed batch size $N$, which is enough for the selected metrics but definitely insufficient for the study of the whole distribution.

\begin{remark}
Please notice that the distribution whose statistics are displayed in the tables is not the one used to maximize \eref{eq:w} since it is a distribution over different starting batches collected with the same random policy and not the distribution that encodes the model uncertainty using the same starting dataset. Indeed, from a bayesian point of view the results are distributed along:
\begin{equation}
    Pr\left(u^{\pi}(M),\mathcal{B}|\pi_{random}\right)=Pr\left(u^{\pi}(M)|\tau_p\right)Pr\left(\tau_p|\mathcal{B}\right)Pr\left(\mathcal{B}|\pi_{random}\right),
\end{equation}
that represents the probability of collecting a batch $\mathcal{B}$ by collecting transitions using a random policy $\pi_{random}$ and hence observing the performance $u^{\pi}(M)$ by deploying a policy $\pi$. Note that there is a deterministic mapping among the posterior $\tau_p$ and the batch, therefore $Pr\left(\tau_p|\mathcal{B}\right)$ is a delta function.
\end{remark}

\subsection{Results and Discussion}
\label{sec:results}

For Ring and Chain, the results averaged over 100 different batches for each batch size $N\in\{8,16,24,32,40,48,56\}$ are displayed in Table \ref{tab:ring-chain-res}.
While for RFL the results averaged over 50 different batches for every batch size $N\in\{15,30,45,60,75,90,105\}$ are reported in Table \ref{tab:fro-res}.

Even if the datasets are composed of relatively short trajectories (Ring and Chain $n=8$ time steps each, Random Frozen Lake $n=15$ time steps each) in most cases UnO does not manage to evaluate the deterministic policies.
Please note that UnO computes the Importance Sampling ratio for a trajectory $h$, a policy $\pi$ and a behavioral policy $\pi_{\beta}$ as
\begin{equation}
\label{eq:unostoch}
\rho_{h}^{\pi} = \prod_{i=1}^{n_h} \dfrac{\pi\left(s_i, a_i\right)}{\pi_{\beta}\left(s_i, a_i\right)}
\end{equation}
where $n_h$ is the number of time steps of the trajectory $h$. However, this formulation assumes that both $\pi$ and $\pi_{\beta}$ are stochastic. In our formulation $\pi_{\beta}\left(s, a\right) = \lvert \mathcal{A}\rvert^{-1}$ $\forall (s,a) \in \mathcal{S} \times \mathcal{A}$, but $\pi$ is stochastic only when it is the output of SPIBB or BOPAH. When $\pi$ is deterministic, the former equation can be rewritten as
\begin{equation}
\label{eq:unodet}
\rho_{h}^{\pi} = \prod_{i=1}^{n_h} \dfrac{\delta_{\pi(s_i), a_i}}{\pi_{\beta}\left(s_i, a_i\right)} = \lvert \mathcal{A} \rvert ^{n_h} \prod_{i=1}^{n_h}  \delta_{\pi(s_i), a_i}.
\end{equation}
This means that $\rho_{h}^{\pi} = \lvert \mathcal{A} \rvert ^{n_h}$ if and only if all sequence of actions and states is consistent with the deterministic policy $\pi$, otherwise $\rho_{h}^{\pi} = 0$. It goes without saying that the probability that the ratio will be zero grows as a power of $\lvert A \rvert$ and exponentially in $n_h$. In particular, the probability that a sequence will be generated by the deterministic policy is in Ring $\lvert A \rvert ^{-n_h} = 3^{-8} \approx 1.5 \times  10^{-4}$, in Chain $2^{-8} \approx 3.9 \times 10^{-3}$ and in RFL $4^{-15} \approx 9.3 \times 10^{-10}$.
Therefore, almost always UnO will pick a policy among SPIBB and BOPAH since the Importance Sampling ratio will be zero for other policies. If even the output of SPIBB and BOPAH will result in a zero Importance Sampling ratio, then the first policy in the candidate set (the trivial policy) will be picked. The said phenomenon is what happens most of the time. Therefore UnO alternates among the Trivial Policy, one among SPIBB and BOPAH, and once in a while it selects another approach.

\paragraph{Ring} Using $q=0.25$ the best method according to the \textit{Max}, \textit{Mean} and the \textit{Median} is NORBU with the \textit{CVaR} Soft Robust objective (see Table \ref{tab:ring-chain-res}). However, the most robust baseline in terms of worst-case performance is BOPAH.
The distributions of results are asymmetric around $\Delta U$. In the cases of BCR and NORBU the \textit{Mean} and the \textit{Median} are approximately zero.
In the cases of SPIBB and BOPAH the \textit{Median} and the \textit{Mean} are less than zero.
Regarding the off-policy evaluation and selection methods, EvC with the \textit{VaR} is the most performing one with respect to all the considered metrics.

\paragraph{Chain} In this environment every baseline except for SPIBB works the same with SPIBB being the worst in terms of \textit{Min} (see Table \ref{tab:ring-chain-res}). Regarding the off-policy evaluation and selection, all algorithms perform well since there is not really a substantial difference between the approaches (except for SPIBB).

\paragraph{Random Frozen Lake (RFL)} We test the approaches in 4 different RFLs. The best approach in terms of overall metrics in 3 environments over 4 is again NORBU with the Soft Robust \textit{CVaR} (see Table \ref{tab:fro-res}). SPIBB is the best in Environment 4. The best selection method is EvC with \textit{VaR}/\textit{CVaR} ($a_q$ / $\phi_q$) which provides identical performances.

\begin{table}[th!]
\caption{Statistics of the normalized performance difference $\Delta U$ between the reported algorithm (risk level $q=0.25$) and the trivial policy averaged over batch size $N\in\{8,16,24,32,40,48,56\}$ with $100$ different batches for size in Ring and Chain. On the right $\Delta U$ with the algorithm selected by EvC and UnO with $a_{0.25}$ and $\phi_{0.25}$. Notice that both EvC and UnO can pick also a policy obtained with a model solved with a different discount factor.}
\label{tab:ring-chain-res}
\resizebox{\linewidth}{!}{%
\begin{tabular}{ll|cccc|cccc}
 &
   &
  \multicolumn{4}{c}{\textbf{Baseline}} &
  \multicolumn{4}{c}{\textbf{Selection Method}}\\
\textbf{Environment} &
  \textit{Metrics} &
  SPIBB &
  BOPAH &
  BCR &
  NORBU &
  $\textit{EvC}_{a_{0.25}}$ &
  $\textit{EvC}_{\phi_{0.25}}$ &
  $\textit{UnO}_{a_{0.25}}$&
  $\textit{UnO}_{\phi_{0.25}}$\\
    \hline
  &
\textit{Max} &
0.61 & 
0.48 & 
0.74 & 
\textbf{0.84} & 
\textbf{0.82} & 
0.71 & 
\textbf{0.82} & 
0.72 \\
\multicolumn{1}{c}{{Ring}} &
\textit{Mean} &
-0.29 & 
-0.28 & 
-0.01 & 
\textbf{0.03} & 
\textbf{0.01} & 
-0.04 & 
-0.26 & 
-0.27 \\
\multicolumn{1}{c}{{}} &
\textit{Median} &
-0.31 & 
-0.34 & 
\textbf{0.0} & 
\textbf{0.0} & 
\textbf{0.0} & 
\textbf{0.0} & 
-0.27 & 
-0.33 \\
\multicolumn{1}{c}{{}} &
\textit{Min} &
-0.78 & 
\textbf{-0.68} & 
-0.82 & 
-0.71 & 
\textbf{-0.82} & 
\textbf{-0.82} & 
-0.96 & 
-0.96\\
\hline
 &
\textit{Max} &
\textbf{0.55} & 
0.54 & 
\textbf{0.55} & 
\textbf{0.55} & 
\textbf{0.55} & 
\textbf{0.55} & 
0.54 & 
0.54 \\
\multicolumn{1}{c}{{Chain}} &
\textit{Mean} &
0.0 & 
0.01 & 
0.01 & 
\textbf{0.02} & 
\textbf{0.01} & 
\textbf{0.01} & 
\textbf{0.01} & 
\textbf{0.01} \\
\multicolumn{1}{c}{{}} &
\textit{Median} &
\textbf{-0.01} & 
\textbf{-0.01} & 
\textbf{-0.01} & 
\textbf{-0.01} & 
\textbf{-0.01} & 
\textbf{-0.01} & 
\textbf{-0.01} & 
\textbf{-0.01} \\
\multicolumn{1}{c}{{}} &
\textit{Min} &
-0.38 & 
-0.16 & 
\textbf{-0.15} & 
\textbf{-0.15} & 
\textbf{-0.16} & 
\textbf{-0.16} & 
\textbf{-0.16} & 
\textbf{-0.16}\\
\hline
\end{tabular}%
}
\end{table}
\begin{table}[th!]
\caption{Statistics of the normalized performance difference $\Delta U$ between the reported algorithm (quantile order used $q=0.25$) and the trivial policy averaged over batch size $N\in\{15,30,45,60,75,90,105,120,135\}$ with $50$ different batches for size in different Random Frozen Lake environments.}
\label{tab:fro-res}
\resizebox{\linewidth}{!}{%
\begin{tabular}{cl|cccc|cccc}
\multicolumn{1}{l}{}                     &                   & \multicolumn{4}{c}{\textbf{Baseline}}                            & \multicolumn{4}{c}{\textbf{Selection Method}}                                                     \\
\multicolumn{1}{l}{\textbf{Environment}} & \textit{Metrics}  & SPIBB          & BOPAH          & BCR           & NORBU          & $\textit{EvC}_{a_{0.25}}$ & $\textit{EvC}_{\phi_{0.25}}$  & $\textit{UnO}_{a_{0.25}}$ & $\textit{UnO}_{\phi_{0.25}}$  \\
\hline

\multicolumn{1}{l}{}                     & \textit{Max}      & \textbf{0.32}  & 0.31           & 0.31          & \textbf{0.32}  & 0.3                   & 0.32                    & \textbf{0.37}         & 0.32                    \\
RFL                                      & \textit{Mean}     & \textbf{0.05}  & -0.04          & -0.04         & \textbf{0.05}  & \textbf{0.05}         & \textbf{\textbf{0.05}}  & -0.02                 & -0.05                   \\
Env. 1                                   & \textit{Median}   & \textbf{0.04}  & -0.07          & -0.04         & \textbf{0.04}  & \textbf{0.04}         & \textbf{\textbf{0.04}}  & -0.01                 & -0.08                   \\
                                         & \textit{Min}      & -0.25          & \textbf{-0.22} & -0.39         & -0.33          & -0.33                 & -0.33                   & -0.31                 & \textbf{-0.22}          \\
\hline
\multicolumn{1}{l}{}                     & \textit{Max}      & 0.33           & 0.22           & 0.3           & \textbf{0.34}  & \textbf{0.34}         & \textbf{\textbf{0.34}}  & 0.28                  & 0.18                    \\
RFL                                      & \textit{Mean}     & 0.02           & -0.07          & -0.05         & \textbf{0.06}  & \textbf{0.06}         & \textbf{\textbf{0.06}}  & 0.0                   & -0.07                   \\
Env. 2                                   & \textit{Median}   & 0.01           & -0.08          & -0.06         & \textbf{0.06}  & \textbf{0.06}         & \textbf{\textbf{0.06}}  & -0.01                 & -0.08                   \\
                                         & \textit{Min}      & -0.21          & -0.22          & -0.29         & \textbf{-0.12} & \textbf{-0.12}        & \textbf{\textbf{-0.12}} & -0.28                 & -0.26                   \\
\hline
\multicolumn{1}{l}{}                     & \textit{Max}      & 0.3            & 0.23           & \textbf{0.43} & 0.36           & \textbf{0.36}         & \textbf{\textbf{0.36}}  & 0.35                  & 0.18                    \\
RFL                                      & \textit{Mean}     & 0.01           & -0.08          & 0.0           & \textbf{0.04}  & \textbf{0.04}         & \textbf{\textbf{0.04}}  & -0.03                 & -0.08                   \\
Env. 3                                   & \textit{Median}   & -0.0           & -0.09          & 0.01          & \textbf{0.02}  & \textbf{0.02}         & \textbf{\textbf{0.02}}  & -0.03                 & -0.09                   \\
                                         & \textit{Min}      & \textbf{-0.16} & -0.3           & -0.36         & -0.27          & -0.27                 & -0.27                   & -0.29                 & \textbf{-0.26}          \\
\hline
\multicolumn{1}{l}{}                     & \textit{Max}      & 0.32           & 0.22           & \textbf{0.36} & 0.31           & \textbf{0.31}         & \textbf{\textbf{0.31}}  & 0.27                  & 0.22                    \\
RFL                                      & \textit{Mean}     & 0.02           & -0.06          & 0.01          & \textbf{0.05}  & \textbf{0.05}         & \textbf{\textbf{0.05}}  & -0.05                 & -0.06                   \\
Env. 4                                   & \textit{Median}   & 0.02           & -0.06          & 0.01          & \textbf{0.05}  & \textbf{0.05}         & \textbf{\textbf{0.05}}  & -0.05                 & -0.06                   \\
                                         & \textit{Min}      & -0.32          & -0.3           & -0.4          & \textbf{-0.29} & \textbf{-0.29}        & \textbf{\textbf{-0.29}} & -0.39                 & -0.3\\
\hline
\end{tabular}%
}
\end{table}

In the following, we comment on the results obtained with EvC. Note this algorithm selects the policy that optimizes the (Conditional) Value at Risk over the first quartile ($q=0.25$) starting from the set of candidate policies discussed in the last section.

\pgfplotstableread{ %
Environment Trivial BCR SPIBB BOPAH NSR Gamma
$a_{0.25}\quad$N=8 17.9 23.2 0.0 0.0 25.5 33.4
$a_{0.25}\quad$N=24 13.9 27.4 0.0 0.0 28.7 30.1
$a_{0.25}\quad$N=40 16.0 27.0 0.0 0.0 25.7 31.4
$a_{0.25}\quad$N=56 16.4 27.0 0.0 0.0 26.5 30.2
$a_{0.25}\quad$N=72 17.7 27.3 0.0 0.0 25.3 29.7
$a_{0.25}\quad$N=88 20.1 27.0 0.0 0.0 24.5 28.5
$\phi_{0.25}\quad$N=8 19.5 25.6 0.0 0.0 17.5 37.5
$\phi_{0.25}\quad$N=24 15.3 28.1 0.0 0.0 24.0 32.6
$\phi_{0.25}\quad$N=40 18.3 25.1 0.0 0.0 22.8 33.9
$\phi_{0.25}\quad$N=56 17.3 26.7 0.0 0.0 24.3 31.7
$\phi_{0.25}\quad$N=72 19.5 25.9 0.0 0.0 23.8 30.8
$\phi_{0.25}\quad$N=88 20.5 26.4 0.0 0.0 23.4 29.7
}\ringdata

\begin{figure}[ht!]
\centering
\begin{tikzpicture}[scale=0.7]

\begin{axis}[
xbar stacked, %
xmin=0, %
ytick=data, %
yticklabels from table={\ringdata}{Environment}, %
legend style={at={(axis cs:115,1)},anchor=south west},
area legend,
every axis plot post/.append style={
solid,
},
]
\addplot[pattern=crosshatch, pattern color=blue] table [x=Trivial, meta=Environment,y expr=\coordindex] {\ringdata};
\addplot[pattern=vertical lines, pattern color=black] table [x=BCR, meta=Environment,y expr=\coordindex] {\ringdata};
\addplot[pattern=dots, pattern color=blue] table [x=SPIBB, meta=Environment,y expr=\coordindex] {\ringdata};
\addplot[pattern=horizontal lines, pattern color=black] table [x=BOPAH, meta=Environment,y expr=\coordindex] {\ringdata};
\addplot[pattern=grid, pattern color=black] table [x=NSR, meta=Environment,y expr=\coordindex] {\ringdata};
\addplot[pattern=checkerboard, pattern color=red!60] table [x=Gamma, meta=Environment,y expr=\coordindex] {\ringdata};

\legend{Trivial,{BCR},{SPIBB},{BOPAH},{NORBU},{Different $\gamma$'s}}
\end{axis}
\end{tikzpicture}
\caption{Policy selection rate by EvC $a_{0.25}$ and EvC $\phi_{0.25}$ in Ring for different batch sizes.}
    \label{fig:bartest-ring}
\end{figure}
\pgfplotstableread{ %
Environment Trivial BCR SPIBB BOPAH NSR Gamma
$a_{0.25}\quad$N=8 1.3 37.3 0.0 0.0 25.8 35.7
$a_{0.25}\quad$N=24 4.8 36.1 0.0 0.0 22.3 36.8
$a_{0.25}\quad$N=40 4.6 35.4 0.0 0.0 23.8 36.3
$a_{0.25}\quad$N=56 6.8 32.5 0.0 0.0 25.1 35.6
$a_{0.25}\quad$N=72 5.5 31.8 0.0 0.0 25.2 37.5
$a_{0.25}\quad$N=88 6.6 32.0 0.0 0.0 24.5 36.8
$\phi_{0.25}\quad$N=8 1.3 37.3 0.0 0.0 25.8 35.7
$\phi_{0.25}\quad$N=24 4.8 35.9 0.0 0.0 22.1 37.1
$\phi_{0.25}\quad$N=40 4.5 35.2 0.0 0.0 23.4 36.8
$\phi_{0.25}\quad$N=56 6.9 31.9 0.0 0.0 24.2 37.0
$\phi_{0.25}\quad$N=72 5.7 31.2 0.0 0.0 24.2 38.9
$\phi_{0.25}\quad$N=88 6.8 31.7 0.0 0.0 24.1 37.5
}\chaindata

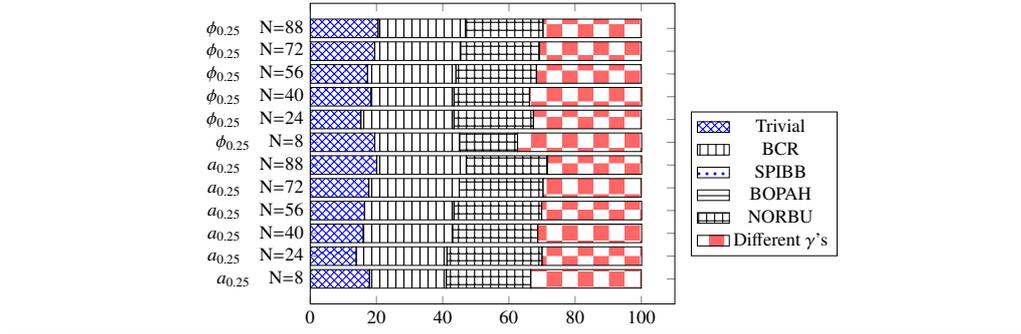
\begin{figure}[ht!]
\centering
\begin{tikzpicture}[scale=0.7]

    \begin{axis}[
xbar stacked, %
xmin=0, %
ytick=data, %
yticklabels from table={\chaindata}{Environment}, %
legend style={at={(axis cs:115,1)},anchor=south west},
area legend,
every axis plot post/.append style={
solid,
},
]
\addplot[pattern=crosshatch, pattern color=blue] table [x=Trivial, meta=Environment,y expr=\coordindex] {\chaindata};
\addplot[pattern=vertical lines, pattern color=black] table [x=BCR, meta=Environment,y expr=\coordindex] {\chaindata};
\addplot[pattern=dots, pattern color=blue] table [x=SPIBB, meta=Environment,y expr=\coordindex] {\chaindata};
\addplot[pattern=horizontal lines, pattern color=black] table [x=BOPAH, meta=Environment,y expr=\coordindex] {\chaindata};
\addplot[pattern=grid, pattern color=black] table [x=NSR, meta=Environment,y expr=\coordindex] {\chaindata};
\addplot[pattern=checkerboard, pattern color=red!60] table [x=Gamma, meta=Environment,y expr=\coordindex] {\chaindata};

\legend{Trivial,{BCR},{SPIBB},{BOPAH},{NORBU},{Different $\gamma$'s}}
\end{axis}
\end{tikzpicture}
\caption{Policy selection rate by EvC $a_{0.25}$ and EvC $\phi_{0.25}$ in Chain for different batch sizes.}
    \label{fig:bartest-chain}
\end{figure}
\pgfplotstableread{ %
Environment Trivial BCR SPIBB BOPAH NSR Gamma
$a_{0.25}\quad$N=15 0.0 0.0 0.0 0.0 40.0 60.0
$a_{0.25}\quad$N=45 0.0 0.0 0.0 0.0 62.7 37.3
$a_{0.25}\quad$N=75 0.0 0.0 0.0 0.0 78.3 21.7
$a_{0.25}\quad$N=105 0.0 0.0 0.0 0.0 90.1 9.9
$a_{0.25}\quad$N=135 0.0 0.0 0.0 0.0 93.9 6.1
$\phi_{0.25}\quad$N=15 0.0 0.0 0.0 0.0 40.0 60.0
$\phi_{0.25}\quad$N=45 0.0 0.0 0.0 0.0 60.0 40.0
$\phi_{0.25}\quad$N=75 0.0 0.0 0.0 0.0 75.2 24.8
$\phi_{0.25}\quad$N=105 0.0 0.0 0.0 0.0 88.6 11.4
$\phi_{0.25}\quad$N=135 0.0 0.0 0.0 0.0 97.6 2.4
}\froaggrodata

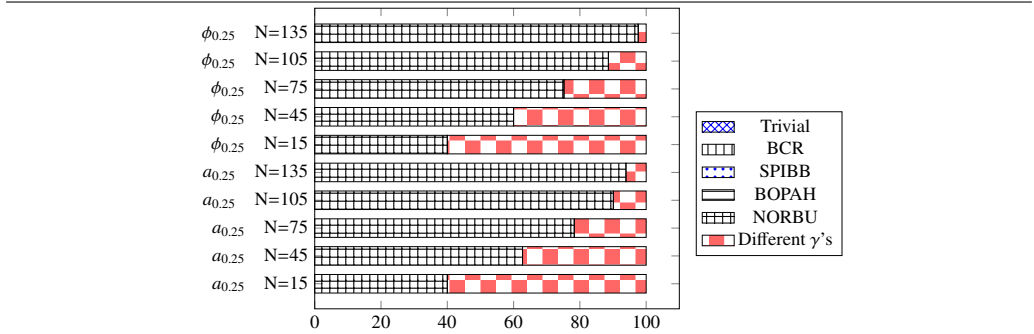
\begin{figure}[ht!]
\centering
\begin{tikzpicture}[scale=0.7]

    \begin{axis}[
xbar stacked, %
xmin=0, %
ytick=data, %
yticklabels from table={\froaggrodata}{Environment}, %
legend style={at={(axis cs:115,1)},anchor=south west},
area legend,
every axis plot post/.append style={
solid,
},
]
\addplot[pattern=crosshatch, pattern color=blue] table [x=Trivial, meta=Environment,y expr=\coordindex] {\froaggrodata};
\addplot[pattern=vertical lines, pattern color=black] table [x=BCR, meta=Environment,y expr=\coordindex] {\froaggrodata};
\addplot[pattern=dots, pattern color=blue] table [x=SPIBB, meta=Environment,y expr=\coordindex] {\froaggrodata};
\addplot[pattern=horizontal lines, pattern color=black] table [x=BOPAH, meta=Environment,y expr=\coordindex] {\froaggrodata};
\addplot[pattern=grid, pattern color=black] table [x=NSR, meta=Environment,y expr=\coordindex] {\froaggrodata};
\addplot[pattern=checkerboard, pattern color=red!60] table [x=Gamma, meta=Environment,y expr=\coordindex] {\froaggrodata};

\legend{Trivial,{BCR},{SPIBB},{BOPAH},{NORBU},{Different $\gamma$'s}}
\end{axis}
\end{tikzpicture}
\caption{Policy selection rate by EvC $a_{0.25}$ and EvC $\phi_{0.25}$ in RFL (aggregate of Env. 1, 2, 3 and 4) for different batch sizes.}
    \label{fig:bartest-aggro}
\end{figure}

In terms of risk awareness, after a global study over different batch sizes, EvC does not select the policy that produces the best values with respect to the considered metrics. Nevertheless, the policy selected by EvC is between the more robust ones. These results are shown in Figures \ref{fig:bartest-ring}, \ref{fig:bartest-chain}, and \ref{fig:bartest-aggro}. In particular, our approach tends to opt for a policy from the ones obtained by solving several models with different discount factors $\gamma$ when the batch is small. The number of times such a policy is selected decreases to the benefit of 1) the trivial policy when the batch size $N$ increases (Ring and Chain) or 2) NORBU (in the RFL environment). This is reasonable since model uncertainty decreases with $N$ and the trivial model will be closer and closer to the true one. We suppose that for not-so-small environments (RFL) the trivial policy can not be trusted for small batch sizes while NORBU manages to cut the posterior space in ambiguity sets that are efficiently optimized over. The policies computed through SPIBB and BOPAH are never selected. Remember that those are stochastic policies that were obtained by improving the batch collector policy that was uniformly random over the actions. Stochastic policies seem not to provide good risk-aware estimates with respect to risk-aware BMDP criteria defined in \eref{eq:w} and also require sampling more models in order, for the method, to estimate a quantile with the needed accuracy.

Another interesting effect reported in Ring is that for $N=8$ the trivial policy is picked a considerable amount of times. Both in Ring and in Chain EvC selects more often the output of BCR rather than that of NORBU, even though NORBU in the end is slightly the most performing according to Table \ref{tab:ring-chain-res}. In RFL only the policies computed through solving different models sampled from the posterior with different $\gamma$'s and NORBU are selected. The first kind of policy is preferred when the batch is very small $N=15$, however, the ratio inverts already for $N=45$ with NORBU that gets more and more chosen with $N$ growing. Both the Trivial policy and the one returned by BCR are always discarded, stressing the superiority of NORBU in this environment typology. Surprisingly, EvC never selects SPIBB nor BOPAH not even in RFL despite its good performance. This is due probably to the difficulty in estimating the quantiles of the performance of a non-deterministic policy such as the output of SPIBB. The algorithm would require a number of sampled models higher than the bail-out hyperparameter.

\section{Conclusion and future work}
\label{sec:conclusions}
This work presents EvC, a method to first evaluate and then select the best risk-aware policies within a set of candidate policies in the context of Offline solutions to Risk-aware Bayesian MDPs.
The Risk-aware BMDP defines an elegant mathematical framework that balances %
the exploitation-caution trade-off in offline model-based sequential decision-making under uncertainty.
The set of candidate policies exploited by EvC contains the strategies obtained by solving not only the trivially learned MDP but also other MDPs with transition dynamics sampled from the Bayesian posterior (e.g. the one shown in \eref{eq:post}) using different discount factors and the solutions of current offline MDP and RL solvers (SPIBB, BOPAH, BCR, NORBU). The estimate of risk in the presented algorithm provides a probabilistic guarantee for the actual performance of the resulting policy described in Theorem \ref{garant} and Theorem \ref{garant2}.
The selected solution maximizes the risk-aware utility function of \eref{eq:w}.
Since EvC is based on the parallel resolution of a great number of models sampled from the Bayesian posterior %
we doubt that it could efficiently scale to select policies for MDPs with a great number of states and actions. However, the presented approach should be considered a valuable tool to be exploited for real-world problem-solving through MDP modeling. In such a case time is an affordable resource since the safety of possible humans in the loop would be the priority.

In the future, we aim to improve EvC's method of generation of the set of candidate policies. An interesting direction consists in %
incrementally enriching the set of candidate policies following some kind of heuristics, e.g. policy improvement by genetic algorithms.
An extension to compute robust policies for data-driven POMDPs could be envisaged whether a consistent representation of the model uncertainty can be formalized.
\section*{Acknowledgments}

This work is supported by the Artificial and Natural Intelligence Toulouse Institute (ANITI) - Institut 3iA (ANR-19-PI3A-0004).

\section*{Code availability}
The code for the experiments is open and available in the Github repository: \url{https://github.com/giorgioangel/evc} %

\bibliographystyle{elsarticle-num}
\bibliography{biblio}

\end{document}